\pdfoutput=1
\documentclass[lettersize,journal]{IEEEtran}
\usepackage{amsmath,amsfonts}
\usepackage{algorithmic}
\usepackage{algorithm}
\usepackage{array}
\usepackage[caption=false,font=normalsize,labelfont=sf,textfont=sf]{subfig}
\usepackage{textcomp}
\usepackage{stfloats}
\usepackage{url}
\usepackage{verbatim}
\usepackage{graphicx}
\usepackage{cite}
\usepackage{diagbox}

\begin{document}

\title{A Transferable Intersection Reconstruction Network for Traffic Speed Prediction}

\author{Pengyu Fu, Liang Chu, Zhuoran Hou, Jincheng Hu, ~\IEEEmembership{Student Member,~IEEE}, Yanjun Huang, ~\IEEEmembership{Member,~IEEE}, and Yuanjian Zhang, ~\IEEEmembership{Member,~IEEE}
\thanks{(Corresponding author: Yuanjian Zhang.)}
\thanks{Pengyu Fu, Liang Chu and Zhuoran Hou are with the Collge of Automotive Engineering, Jilin University, Changchun 130022, China (e-mail: fupy20@mails.jlu.edu.cn; chuliang@jlu.edu.cn; houzr20@mails.jlu.edu.cn).}

\thanks{Jincheng Hu and Yuanjian Zhang are with the     Department of Aeronautical and Automotive Engineering, Loughborough University, Loughborough, U.K (e-mail: jincheng.hu2020@outlook.com; y.y.zhang@lboro.ac.uk).}

\thanks{Yanjun Huang is with the School of Automotive Studies, Tongji University, Shanghai, China. (e-mail:huangyanjun404@gmail.com).}}

\markboth{Journal of \LaTeX\ Class Files,~Vol.~14, No.~8, August~2021}%
{Shell \MakeLowercase{\textit{et al.}}: A Sample Article Using IEEEtran.cls for IEEE Journals}

\IEEEpubid{0000--0000/00\$00.00~\copyright~2021 IEEE}

\maketitle

\begin{abstract}
Traffic speed prediction is the key to many valuable applications, and it is also a challenging task because of its various influencing factors. Recent work attempts to obtain more information through various hybrid models, thereby improving the prediction accuracy. However, the spatial information acquisition schemes of these methods have two-level differentiation problems. Either the modeling is simple but contains little spatial information, or the modeling is complete but lacks flexibility. In order to introduce more spatial information on the basis of ensuring flexibility, this paper proposes IRNet (Transferable Intersection Reconstruction Network). First, this paper reconstructs the intersection into a virtual intersection with the same structure, which simplifies the topology of the road network. Then, the spatial information is subdivided into intersection information and sequence information of traffic flow direction, and spatiotemporal features are obtained through various models. Third, a self-attention mechanism is used to fuse spatiotemporal features for prediction. In the comparison experiment with the baseline, not only the prediction effect, but also the transfer performance has obvious advantages.
\end{abstract}


\section{Introduction}
\IEEEPARstart{W}{ith} the development of the automobile industry and the improvement of people's living standards, the number of automobiles has continued to increase. At the same time, the construction speed of infrastructure lags behind the growth rate of vehicles, which aggravates the problems of traffic congestion and low transportation efficiency. Road traffic speed is the average speed of vehicles on the road, and it is an important indicator for evaluating traffic conditions. Improving the utilization efficiency of infrastructure by predicting Road traffic speed is a hot spot to solve traffic problems at present. For example, drivers can avoid slow road and choose better driving routes based on predicted road traffic speeds. On-board computers can adjust energy management strategies based on predicted road traffic speeds to save energy and reduce emissions.Traffic managers can manage road networks and allocate resources systematically based on predicted road traffic speeds \cite{zhang1,zhang2,1,2}. 

Since the road traffic speed is closely related to the driving state of the vehicles on the road, road traffic speed is mainly affected by two aspects. (1) The impact of historical road traffic speeds. Although the road traffic speed changes all the time, it is impossible to completely separate from the historical road speed. (2) The impact of the traffic speed on adjacent roads. In a road network, different roads are connected by intersections. When a road is congested, it will directly affect the entry and exit of vehicles on adjacent roads. Therefore, the prediction of road traffic speed is a challenging task.

In the current researches, traffic speed prediction methods are mainly divided into two categories: model-driven methods \cite{14,15,16,17} that rely on theoretical mathematical models and data-driven methods that use statistical methods \cite{26,27,5} and artificial intelligence algorithms \cite{7,30,31}. Model-driven approaches use prior knowledge and physical properties to model traffic flow and predict road traffic speeds through traffic simulation. Therefore, the model-driven approach is highly interpretable. At the same time, less historical data is required when building the model \cite{4}. However, model-driven methods require a large number of assumptions, such as time-invariant travel costs and homogeneous traveler route choices \cite{2017Spatiotemporal}. The accuracy of the model-driven approach depends on how accurately the model is built. However, in the process of building a high-precision model, different assumptions need to be adopted for different traffic conditions, which leads to an overly complex model.

Data-driven methods are data-dependent techniques that can break through the shackles of model-driven methods, and their effectiveness largely depends on the quality and size of traffic datasets. As data acquisition and storage become less difficult, data-driven approaches such as statistical methods and artificial intelligence algorithms are becoming more popular. In the early research works, data-driven methods for traffic speed prediction generally took the historical speed sequence of the target road as input, and found the change rule of traffic speed to predict the future traffic speed. Including statistical based methods such as ARIMA (Auto Regressive Integrated Moving Average Model) \cite{5}, KF (Kalman Filter Theory) \cite{6}, artificial intelligence algorithms such as machine learning algorithm  SVR (Support Vector Regression) \cite{7}, and deep learning algorithm LSTM (Long and Short-term Memory) \cite{8,82} and so on. 
\IEEEpubidadjcol
However, the above data-driven methods only focus on the historical data of the target road while the historical data of the adjacent roads are not utilized. To solve this problem, some works flattened the topology of the road network and add temporal information, so that the spatiotemporal information of traffic is converted into 2D images. In this way, the temporal and spatial correlations could be simultaneously obtained through CNN (convolutional neural network) \cite{9}. However, due to the characteristics of CNN, compared with RNN (Recurrent Neural Network), which specializes in processing sequence data, CNN is relatively limited in its ability to mine time series information in data. This has prompted people to explore the fusion of multiple models. Some works used CNN to obtain spatial correlation and RNN to obtain temporal correlation, and obtain better prediction results \cite{8489600}. Nevertheless, the transformed 2D image is difficult to reflect the topology of the real road network. Especially when the number of adjacent roads increases and the road network structure is complex, this processing method will produce huge deviations. And GCN (Graph Convolutional Network) shown excellent performance in spatial modeling of road networks in several recent works \cite{10,11,12}. The limitation of GCN is that it requires full-batch training, so it takes up a lot of GPU memory during calculation. At the same time, GCN can only predict the speed of the road that participated in the training. Once the graph structure changes, the entire model needs to be rebuilt and retrained.

In general, existing traffic speed prediction methods tend to use flexible data-driven methods. The temporal and spatial features are extracted through a deep learning hybrid model to obtain higher prediction accuracy. Spatial feature extraction methods present a state of two-level differentiation. One is that all adjacent roads are tiled and each road can only be adjacent to two roads, which is far from the actual road structure and only retains part of the spatial information. The other is to consider the relationship between all of the roads, which brings huge computational pressure. At the same time, these two spatial feature extraction methods have the same problem. After the road network structure changes, the entire spatial feature extraction module needs to be redesigned and retrained, so the model has poor generality. This paper proposes IRNet (Transferable Intersection Reconstruction Network). The model defines a new method for obtaining spatial information of roads and combines spatiotemporal features to predict road traffic speed. The work and contributions of this paper are as follows:
\begin{enumerate}
\item{A new multi-model fusion deep learning architecture is proposed, which simultaneously possesses good predictive performance and transfer ability. It is mainly divided into three modules according to functions. The data generation module obtains spatiotemporal information. The feature extraction module obtains spatiotemporal features through multiple deep learning models. The feature fusion module fuses multiple spatiotemporal features to predict the road traffic speed.}
\item{A new spatial feature extraction method called intersection reconstruction is defined. It simplifies the complex road network topology into the same road network structure. Therefore, when the road network structure changes, the prediction model does not need to change the structure. The expansion ability and transfer ability of the model are improved.}
\item{CNN has the characteristics of weight sharing and local connection, which is very suitable for extracting the spatial features of virtual intersections. LSTM is suitable for extracting spatial sequence features of traffic flow direction. By combining CNN and LSTM, the utilization of spatial information is improved, and highly representative spatial features can be learned.}
\item{The self-attention mechanism is introduced into the feature fusion module, which reflects the influence of different spatiotemporal features on the prediction of target road traffic speed, and is highly interpretable.}
\end{enumerate}

The rest of this paper is organized as follows. Section \uppercase\expandafter{\romannumeral2} reviews related work. Section \uppercase\expandafter{\romannumeral3} introduces some preliminary knowledge. Section \uppercase\expandafter{\romannumeral4} introduces the method of model building in detail. Section \uppercase\expandafter{\romannumeral5} reports the experimental results. Section \uppercase\expandafter{\romannumeral6} concludes the paper.

\section{Related work}
In model-based traffic speed prediction, methods using traffic flow theory are the focus of research. Traffic flow theory is based on mathematical and physical methods such as probability theory, mathematical statistics and calculus \cite{13}. According to different research subjects, it can be mainly divided into microscopic models and macroscopic models. The research subject of the microscopic models is the vehicles on the road network, and the queuing phenomenon of the vehicles is evolved through the running process of the vehicles on the road network. The representative methods are the car-following model \cite{14,15} and the cellular automaton model \cite{16,17}. According to the theories of fluid dynamics and aerodynamics, the macroscopic models take the traffic flow as the main research body to establish the traffic model and solve the traffic evolution law. The representative methods are the velocity-density model \cite{18} and the gas dynamics model \cite{19}. In addition, the microscopic model and the macroscopic model are linked through the traffic wave theory, and a traffic flow model between microscopic and macroscopic is established. This model has common features of both microscopic model and macroscopic model \cite{20}, and its purpose is to combine the advantages of both while overcoming their respective limitations.

The derivation process of the traffic flow model is more rigorous, and the physical meaning of the model is clear. But some simplifying assumptions are required \cite{21}. It also leads to the result that the traffic speed prediction accuracy of the model-based method largely depends on the accuracy of the traffic flow model. In order to improve the accuracy of the traffic flow model, Zheng et al. \cite{22} established a driver's emergency response mechanism based on the traffic flow model. Gao et al. \cite{23} focused on the impact of low-visibility weather on car-following driving behavior. Geng et al. \cite{24} studied the car-following driving behavior of different vehicle types under different environmental conditions. The addition of mechanisms can improve the accuracy of the traffic flow model, and with it comes the complexity of the building process. Furthermore, it is very complicated to add additional measurement data after the model is established \cite{2015Towards}.

With the development of computer computing power and the development of various data storage devices, The data-driven approach has become a hot spot. Data-driven methods used in the field of traffic forecasting are mainly divided into statistical methods and artificial intelligence algorithms, and artificial intelligence algorithms can be subdivided into basic machine learning algorithms and deep learning algorithms. The statistical methods are to obtain the relationship between the variables through mathematical statistics method after the data is obtained. For traffic forecasting, it is usually assumed that the population obeys a certain distribution, and this distribution can be determined by some parameters. The model constructed on this basis is also called a parametric model. Such as HA (Historical Average Model) \cite{26}, AR (Auto Regressive Model) \cite{AR} and MA (Moving Average Model) \cite{27}. With the development of the study, a number of more complex prediction methods with higher accuracy have emerged, including Kalman Filtering Model \cite{6}, Exponential Smoothing Model \cite{28}, etc. Among them, ARIMA (Autoregressive Integrated Moving Average) considering the periodicity of speed data has a good effect in predicting traffic speed \cite{5}. ARIMA has good theoretical interpretability and clear calculation structure, and is suitable for the prediction of road traffic speed. However, statistical methods fit data in a fixed functional form, usually only for simple problems. And the parameter tuning process is too complicated, resulting in poor online performance.

Compared with statistical methods, artificial intelligence algorithms can fit a variety of different functional forms and are more flexible to use. At the same time, the artificial intelligence algorithms have few prior assumptions about the input variables. Therefore, these methods have good fitting effect, and are more suitable for dealing with missing values and noisy data. However, artificial intelligence algorithms require more training data for fitting the objective function, and there is a higher risk of overfitting. Basic machine learning algorithms include SVR (Support Vector Regression) \cite{7,30}, RF (Random Forest algorithm) \cite{31}, KNN (k-neighbor algorithm) \cite{32}, Bayesian model \cite{33} and so on. The disadvantage of basic machine learning algorithms is that they lack the ability to analyze large-scale data. Therefore, the basic machine learning algorithm needs to select features manually, and the selection of features directly affects the prediction effect. Correspondingly, deep learning methods using two or more hidden layers can capture the feature expression in the input data through a multi-layer structure.Compared with the method of constructing features by artificial rules, deep learning algorithms can better describe the intrinsic information of data. In recent years, a great deal of literature has been published on short-term future speed prediction using different kinds of deep learning methods, such as BPNN (Backpropagation Neural Networks) \cite{34}, DBN (Deep Belief Networks) \cite{37} and SAEs (Stacked Autoencoders) \cite{38}.

In deep learning algorithms, RNN (Recurrent Neural Network) can use internal storage units to process input sequences, thus giving neural networks the ability to learn time series. Due to the dynamic nature of traffic systems, RNN is particularly suitable for capturing the temporal evolution of traffic flow. LSTM (Long Short-Term Memory Network) is variants of RNN and have solved the vanishing gradient problem of RNN \cite{9}. Ma et al. \cite{39} used LSTM to capture long-term temporal dependencies in short-term vehicle speed prediction, and could determine the optimal time window for time series. Zhao et al. \cite{40} integrated the origin destination correlation (ODC) in the LSTM network through a fully connected layer and a vector generator. However, the fully connected layer directly extracts the spatiotemporal features of all road networks, and does not focus on the target road that needs to be predicted at the end, and the spatial structure is not expressive enough. As a classic algorithm of computer vision, CNN (Convolutional Neural Network) is very good at capturing spatial information. Ma et al. \cite{9} transformed the spatiotemporal information of traffic flow into 2D images according to the horizontal axis as time and the vertical axis as each road. and then obtained the spatiotemporal correlation of traffic flow simultaneously through CNN. Since CNN and LSTM have their own advantages in capturing spatiotemporal features, Wu et al. \cite{41} proposed to obtain the spatial and temporal features of traffic flow through CNN and LSTM respectively, and finally combined all features for traffic flow prediction. However, CNN and LSTM do not interfere with each other, and the feature fusion is insufficient. Liu et al. \cite{42} further refined the acquisition process of spatiotemporal features. The spatial features are captured by CNN, and then the captured spatial features are input into LSTM to capture the temporal features. However, there is a problem with the above methods. In the converted 2D image, the time axis is the same as the real situation, but the spatial axis is forced to flatten the complex spatial information. It does not reflect the topology of the road network, especially when the road network is complex.

GCN (Graph Convolutional Network), a feature extractor similar to CNN, is a deep learning network based on graph theory. The object of GCN is the graph describing non-European relations. Zhao et al. \cite{10} used GCN to capture spatial features, and then used GRU to obtain temporal features, and fused the features to output prediction results. Lv et al. \cite{12} introduced semantic information, encoded the correlation between roads through multiple graphs. Multiple GCNs were fused to capture spatial features, and then GRU was used to capture temporal features to obtain prediction results. GCN models the entire road network to obtain complete spatial information. But the disadvantage is that GCN is full-batch training, so it is difficult to apply to a large-scale road network. In the same way, GCN has poor flexibility and cannot deal with unseen nodes at all. When predicting the traffic speed of a road not involved in training, the graph needs to be restructured and the model needs to be retrained.
\section{Preliminary}
\subsection{Graph Theory}
In graph theory, the graph is composed of vertices and edges connecting two vertices. This kind of graph usually uses vertices to represent objects, and an edge connecting two vertices indicates the relationship between the corresponding two objects. The graph structure can effectively model and solve road network problems that need to consider the relationship between roads.

The road network can be defined as a graph $G=\left(V,E\right)$, where $V=\left\{v_1,v_2,\ldots,v_i\right\}$ is the set of vertices and each vertex $v_i$ represents a road $r_i$, $E\subseteq\left\{e_{ij}=v_iv_j\mid{v_i,v_j\in{V}}\right\}$ is the set of edges and each edge $e_{ij}$ represents the correlation between $v_i$ and $v_j$. In an undirected graph, $v_iv_j=v_jv_i$. For road vertex $v_i$, if there is an edge $e_{ij}\in{E}$, road vertex $v_j$ is a neighbor of road vertex $v_i$, denoted as $v_j\in{N_i}$. In other words, traffic flow on road $r_i$ can flow to road $r_j$ through the intersection. Let $\left\|V\right\|=n$, the graph can be represented by an adjacency matrix $A\in{\mathbb{R}^{n\times{n}}}$. The value of $A_{i,j}$ reflects the adjacency relationship between $v_i$ and $v_j$ as follows:

\begin{equation}
\label{eq1}
A_{i,j}=\begin{cases}
1&e_{ij}\in{E}
\\0&e_{ij}\notin{E}
\end{cases}
\end{equation}
\subsection{Graph of Road}
In this paper, the road is regarded as the vertex of the graph, the intersection connecting the road is the edge of the graph, and the value of the edge indicates the connection between the vertices. The average speed of the road is the feature of the vertex, and the average speeds of dual lanes of the same road are not the same, so this paper treats the dual lanes of the same road as two independent roads. Fig. 1(a) gives an example of a simplified road network and the roads in the figure are all one-way streets in the direction of the arrow. According to the description of graph theory in 3.1, the directed graph shown in Fig. 1(b) can be constructed by considering the roads as vertices. 

\begin{figure}[!t]
\centering
\subfloat[]{\includegraphics[width=1.8in]{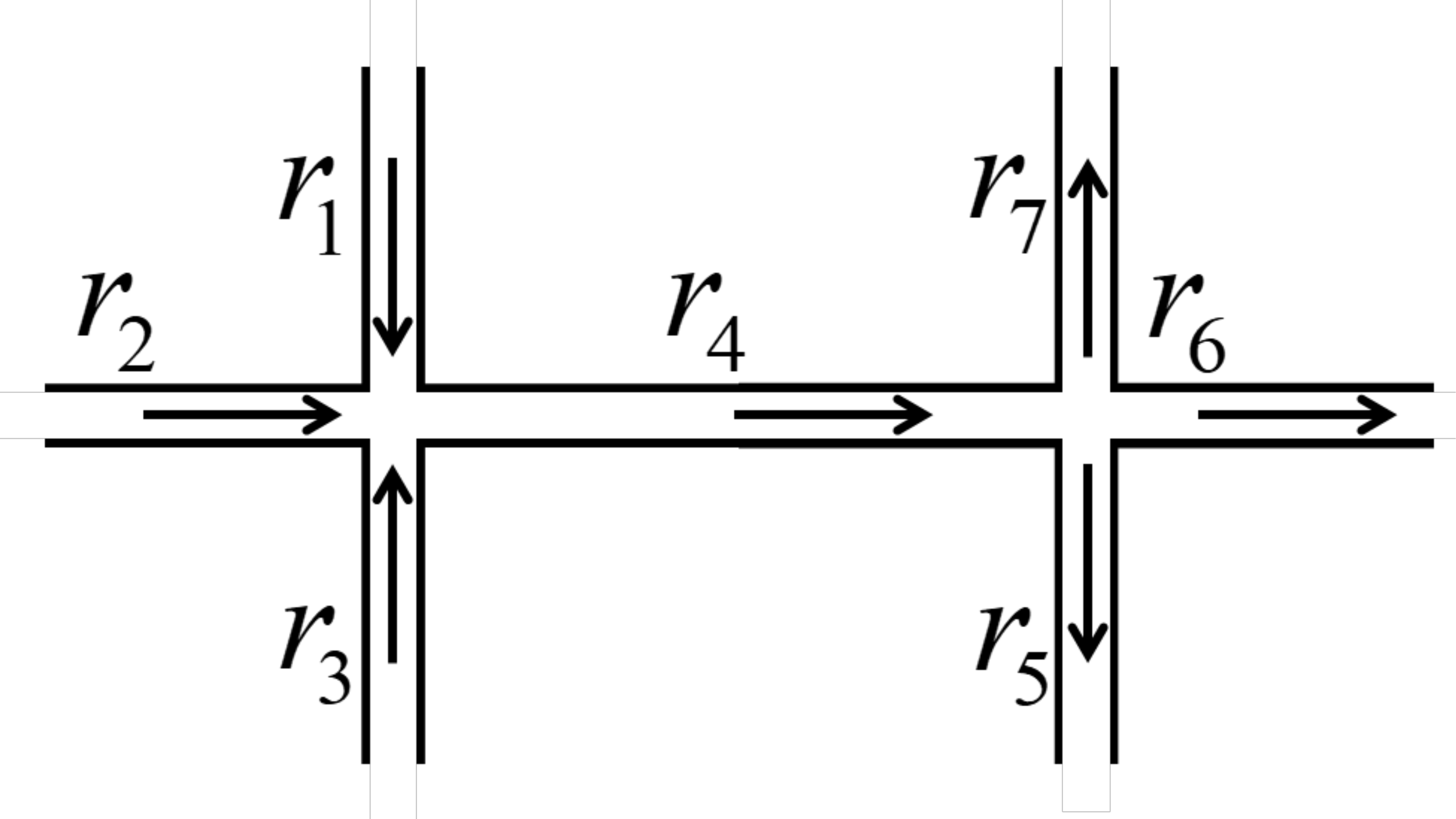}%
\label{fig1a}}
\hfil
\subfloat[]{\includegraphics[width=1.6in]{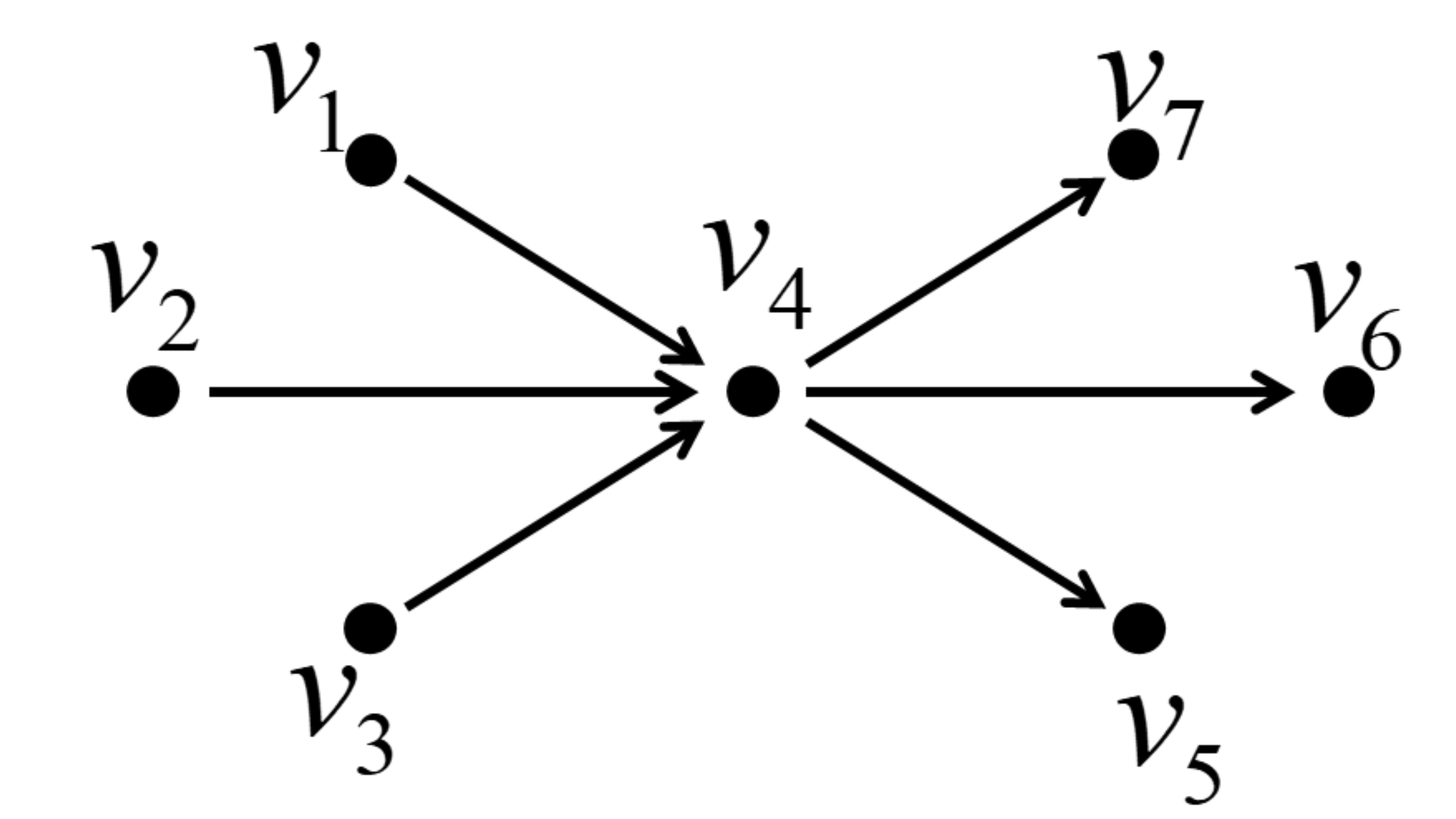}%
\label{fig1b}}
\caption{(a) an example of a simplified road network.(b) A directed graph constructed for the road structure shown in (a).}
\label{fig_1}
\end{figure}

According to the traffic flow theory, the vehicles on the road can be regarded as fluids and flow in the direction specified by the road. $r_i\to{r_j}$ means that vehicles on $r_i$ can merge into $r_j$ through the intersection. If $r_i\to{r_j}$, $r_j$ is the downstream adjacent road of $r_i$, denoted as $r_j\in{DR_i}$. Conversely, $r_i$ is the upstream adjacent road of $r_j$, denoted as $r_i\in{UR_j}$. In addition, this paper defines that if $r_j\in{DR_i}$ and $r_k\in{DR_j}$, $r_k$ is the 2-order downstream adjacent road of $r_i$, denoted as $r_k\in{DR_i^2}$, and so on. Similarly, $r_i$ is the 2-order upstream adjacent road of $r_k$, denoted as $r_i\in{UR_k^2}$. For the example in Fig. 1(b), $r_6\in{DR_4^1}$, $r_2\in{UR_6^2}$.
\subsection{Problem}
The study in this paper is to predict the average traffic speed of road nodes. There are three relevant parameters introduced here:

\begin{enumerate}
\item{The length of historical sequence $h$. $h$ is an important parameter to measure the amount of time information. The average speed of the road $r_i$ at time $t$ is recorded as $speed_i^t$, then $S_i^t=\left[speed_i^{t-h+1},\ldots,speed_i^t\right]$ is historical data sequence of the road $r_i$ at time $t$ of length $h$.}
\item{Space width $w$. $w$ is an important parameter to measure the amount of spatial information. Space width in this paper refers to the order of adjacent roads centered on the target road, rather than the distance of the physical space.}
\item{Prediction horizon $p$. $p$ is a parameter to measure the predictive performance of the model, representing the span of the moment we predict based on the existing data. The average traffic speed of the target road with a prediction span of $p$ is recorded as $speed_{tar}^{t+p}$.}
\end{enumerate}

According to the above parameters, we define the problem as follows: For target road $r_{tar}$, find all adjacent roads with the space width $w$. At time $t$, $speed_{tar}^{t+p}$ is predicted according to the historical speed sequence of the adjacent roads and the target road with a length of $t$.
\section{IRNet method}
The model as shown in Fig. 2 is divided into three modules according to functions, the data generation module, the feature extraction module and the feature fusion module. The function of the data generation module is to generate input data from all the original road historical data according to the preset input parameters. The feature extraction module extracts the intersection features of each intersection through the convolution layer, and then extracts the temporal and spatial features of the adjacent roads through the Temporal-LSTM and Spatial-LSTM. The feature fusion module uses the self-attention mechanism to quantify the influence of spatio-temporal features of different orders on the final result and then outputs the prediction result.

\begin{figure}[!t]
\centering
\includegraphics[width=3.4in]{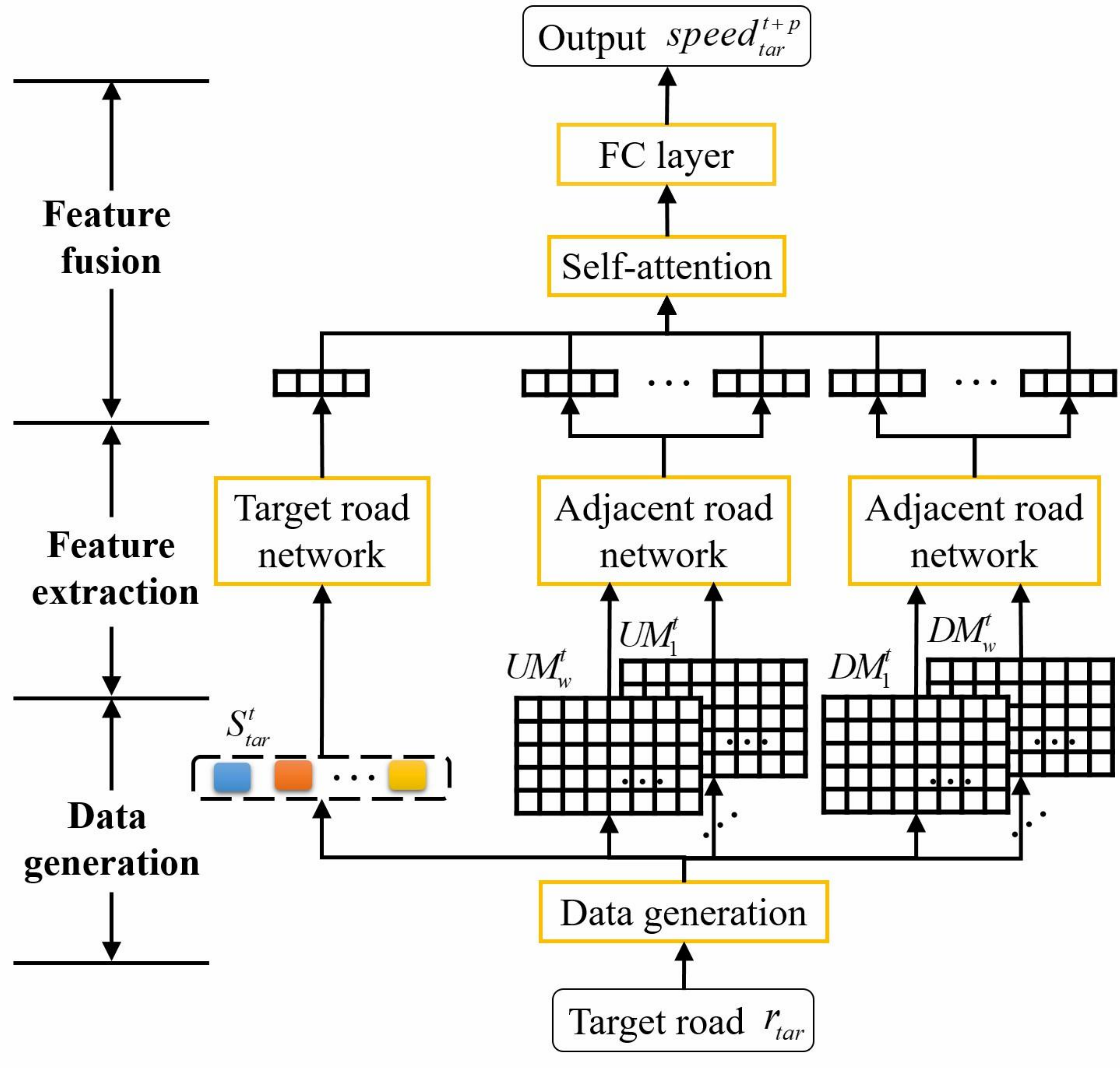}
\caption{The overall architecture of IRNet.}
\label{fig2}
\end{figure}

\subsection{Data Generation Module}
The data generation module, as shown in Fig. 3(a), generates the input feature matrix of the target road network and the adjacent road networks. The key step in the generation of the input data of the adjacent road network is to simplify the complex road network topology into the intersection reconstruction operation of the same road network structure.

$S_{tar}^t=\left[speed_{tar}^{t-h+1},\ldots,speed_{tar}^t\right]$ is the input data of the target road network, which is the historical speed sequence of the target road $r_{tar}$ at time   with a length of $h$.

The input data of the adjacent road network is generated in two parts. Firstly, the complex road network topology is simplified through the intersection reconstruction operation, and the ordered sets of adjacent roads are obtained. Then, the sorted sets are transformed into the input data of the adjacent road network.

\begin{figure}[!t]
\centering
\subfloat[]{\includegraphics[width=2.2in]{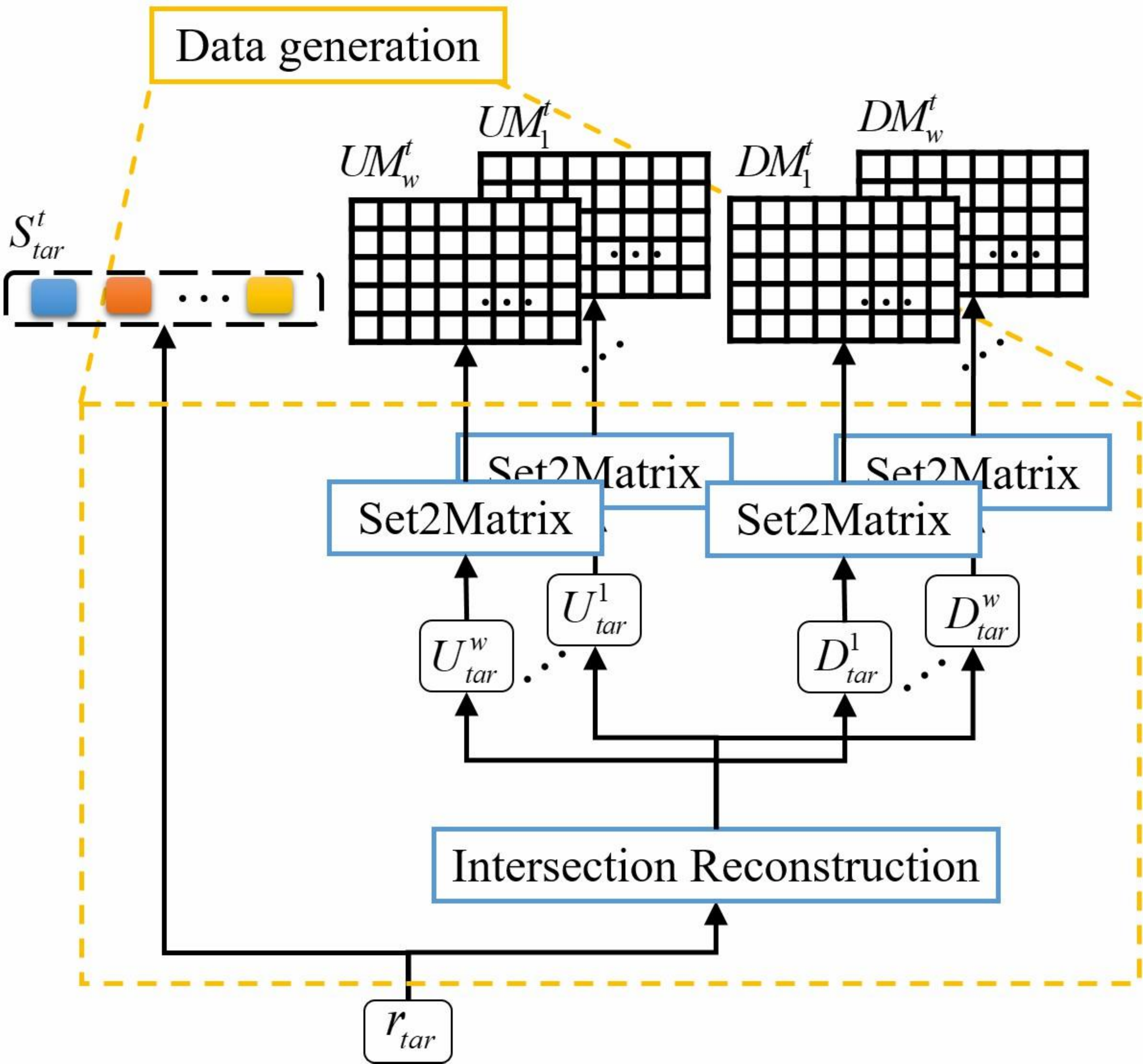}%
\label{fig3a}}
\hfil
\subfloat[]{\includegraphics[width=1.2in]{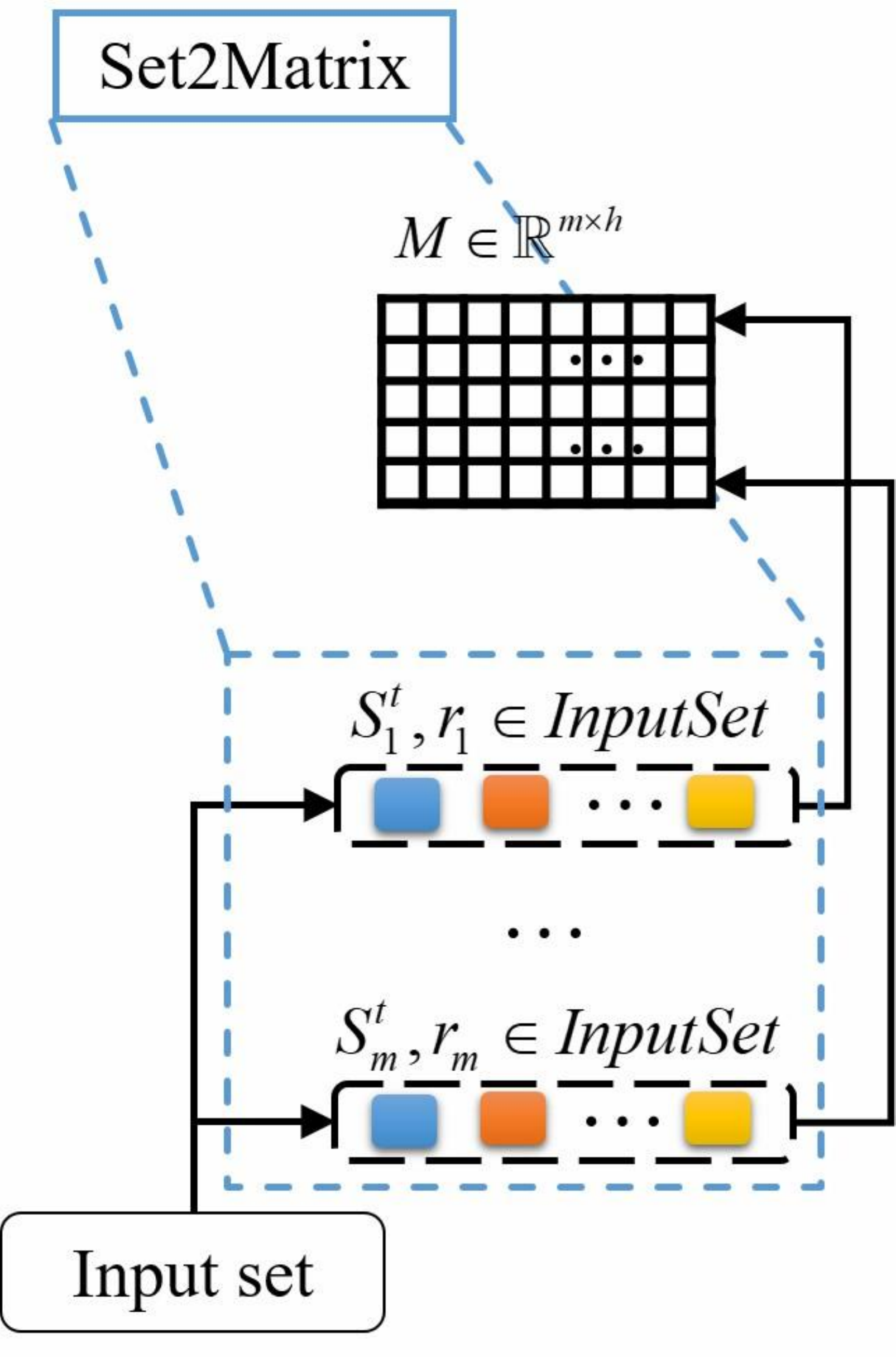}%
\label{fig3b}}
\caption{(a) The structure of the data generation module. (b) The process of set to matrix.}
\label{fig_3}
\end{figure}

\subsubsection{Intersection Reconstruction}
There are many kinds of intersections in reality, as shown in Fig. 4. In most cases, the number of adjacent roads of different target roads is different. This paper proposes a method to reconstruction a variety of intersections into a class of intersection with the same structure without changing the characteristics of the original intersection as much as possible. The process of intersection reconstruction is divided into three steps. First, the roads at the intersection are numbered and sorted to ensure the similarity of intersection functions. Then normalize the number of roads connected to each intersection to ensure the similarity of the intersection structure. Finally, in order to reflect the road network structure after reconstructing the intersection, the sorting method of high-order adjacent roads is defined.

\begin{figure}[!t]
\centering
\includegraphics[width=2.5in]{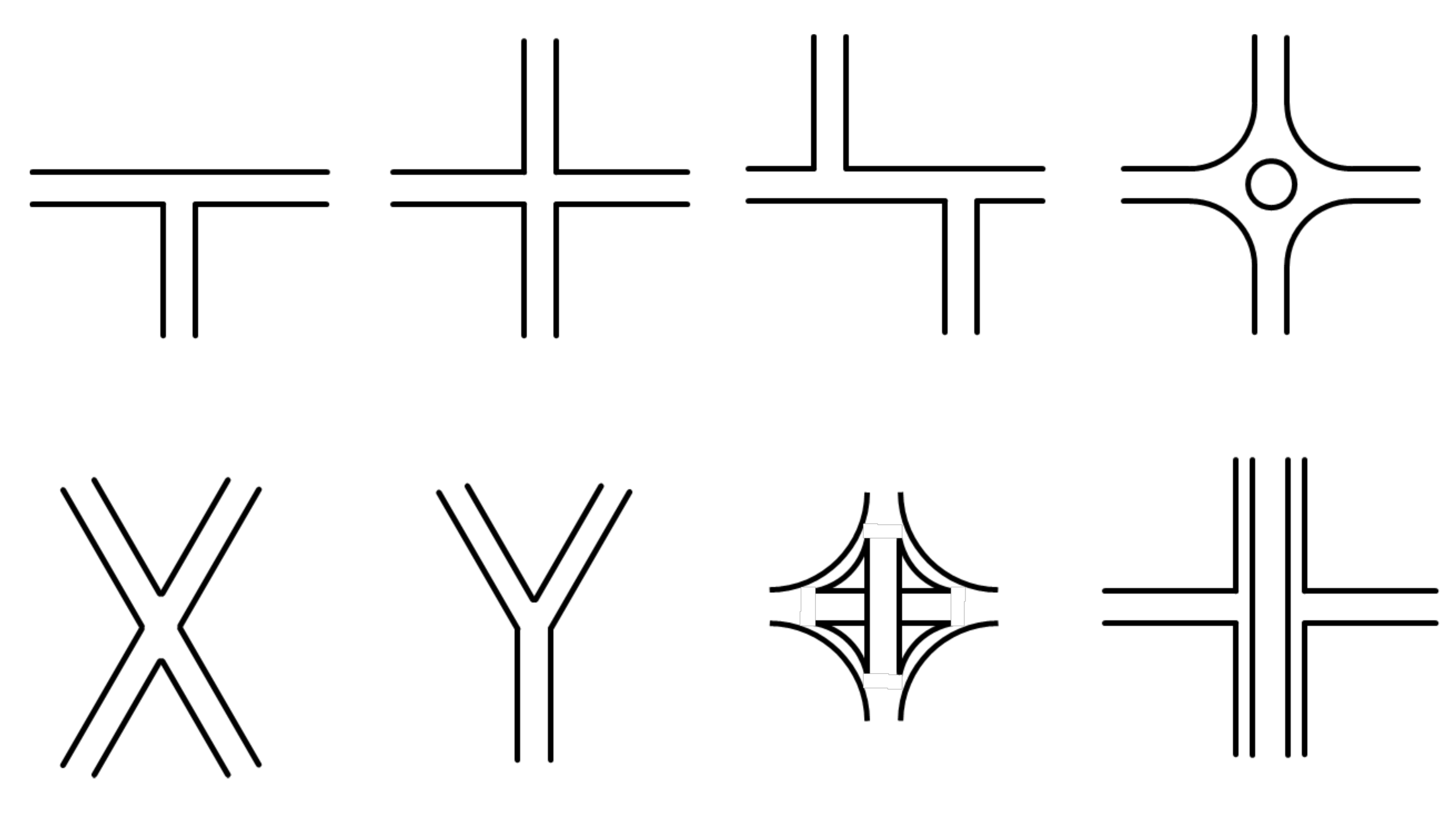}
\caption{Various intersections in the real road network.}
\label{fig4}
\end{figure}

Since the functions and structures of upstream and downstream intersections are completely different, all upstream intersections should be reconstructed into one type of intersection and all downstream intersections should be reconstructed into another type of intersection. Here, the upstream intersection is used as an example to introduce how to reconstruct, and the downstream intersection is similar.
\paragraph{Sort The Adjacent Roads}
This paper studies the prediction of traffic speed, focusing on the impact of adjacent roads on the future traffic speed of the target road. Therefore, a historical road speed sequence is taken as the feature vector of the road to reflect the speed characteristics of the road. The feature vector of road vertex $v_i$ is $F_i=\left[speed_i^1,\ldots,speed_i^\tau\right]$, where $speed_i^t$ is the road speed of road vertex $v_i$ at time $t$, and $\tau$ is the length of the historical speed sequence. After obtaining the feature vectors of all roads, the correlation of any two feature vectors can be calculated. When studying the speed of the road, the correlation of the road feature vector is approximately equal to the correlation of the road speed.

This paper uses DTW (Dynamic Time Warping algorithm) \cite{43} to calculate the correlation among feature vectors. Compared with the Euclidean distance calculated by corresponding points at the same position of the two feature vectors, DTW can map a point of one feature vector to multiple points of another feature vector. In other words, any point of the feature vector will find the point on the other feature vector with the smallest possible distance to calculate the distance. In order to evaluate the impact of the speed of the adjacent road on the speed of the target road, the similarity of the overall trend is more important than the similarity of each moment. DTW has the effect of stretching or compressing the sequence, so it is more suitable for measuring the correlation of the road feature vectors. The formula is as follows:
\begin{equation}
\begin{aligned}
\label{eq2}
&DTW\left(A_i,B_j\right)=\\
&\sqrt[q]{\left|a_i-b_j\right|^q+\left(min\left\{\begin{matrix} DTW\left(A_{i-1},B_{j-1}\right) \\ 
DTW\left(A_{i},B_{j-1}\right)\\
DTW\left(A_{i-1},B_{j}\right)\end{matrix}\right\}\right)^q}
\end{aligned}
\end{equation}
where $A_i$ is the sequence $\left(a_1,\ldots,a_i\right)$, $\left|a_i-b_j\right|^q$ is the  $q$-order distance between the $i$th point of $A_i$ and the $j$th point of $B_j$, Let $q=2$ in this paper.

For any target road, sort the upstream adjacent roads according to Algorithm 1. It can be seen that the label of the adjacent road indicates the degree of correlation between the adjacent road and the target road. Adjacent roads with the same label number of different target roads play the similar structural roles in their respective intersections. 
\begin{algorithm}[H]
\caption{Sort the upstream adjacent roads of the target road ($SORTUP$)}\label{alg:alg1}
\begin{algorithmic}[1]
\REQUIRE Target road $r_{tar}$; The set of 1-order upstream adjacent roads of the target road $UR_{tar}^1$. 
\ENSURE The ordered set of upstream adjacent roads of the target road $SUR_{tar}$. 
\STATE \hspace{0.5cm}Initializing $z\gets{0}$;
\STATE \hspace{0.5cm}\textbf{while} $z<\left|UR_{tar}^1\right|$ \textbf{do}
\STATE \hspace{1cm}$z\gets{z+1}$
\STATE \hspace{1cm}Finding the $z$th road $r_z$ in the set of 1-order\\
\hspace{1cm}upstream adjacent roads, $r_z=UR_{tar}^1[z]$;
\STATE \hspace{1cm}Calculating the correlation between $r_z$ and $r_{tar}$,\\
\hspace{1cm}and take the result as the label $l_z$ of $r_z$,\\ \hspace{1cm}$l_z\gets{DTW\left(F_i,F_{tar}\right)}$.
\STATE \hspace{0.5cm}Sorting the roads from small to large according to the \\
\hspace{0.5cm}label to get the sorted set of upstream adjacent roads \\
\hspace{0.5cm}of the target road $SUR_{tar}$.
\STATE \hspace{0.5cm}\textbf{return}  $SUR_{tar}$
\end{algorithmic}
\label{alg1}
\end{algorithm}
\paragraph{Normalization of Intersection}
By sorting the adjacent roads, various real intersections are abstracted into similar virtual intersections. The structures of these intersections are similar, but the number of connecting roads at each intersection is probably different. This paper stipulates that each intersection is fixedly connected $k$ adjacent roads to the target road. All virtual intersections are normalized according to Algorithm 2, and reconstructed intersections with completely consistent structure are obtained.
\begin{algorithm}[H]
\caption{Normalization of upstream adjacent roads of the target road($NORMUP$)}\label{alg:alg2}
\begin{algorithmic}[1]
\REQUIRE Target road $r_{tar}$; The set of 1-order upstream adjacent roads of the target road $UR_{tar}^1$;The normative number $k$ of adjacent roads of the target road. 
\ENSURE The normalized set of upstream adjacent roads of the target road $NUR_{tar}$. 
\STATE \hspace{0.5cm}Sorting the upstream adjacent roads of the target\\
\hspace{0.5cm}road $SUR_{tar}=SORTUP\left(r_{tar},UR_{tar}^1\right)$;
\STATE \hspace{0.5cm}\textbf{if} $\left|SUR_{tar}\right|>k$ \textbf{then}
\STATE \hspace{1cm}$NUR_{tar}\gets{}$ the first $k$ roads in $SUR_{tar}$
\STATE \hspace{0.5cm}\textbf{else if} $\left|SUR_{tar}\right|=k$ \textbf{then}
\STATE \hspace{1cm}$NUR_{tar}\gets{SUR_{tar}}$
\STATE \hspace{0.5cm}\textbf{else}
\STATE \hspace{1cm}$NUR_{tar}\gets{SUR_{tar}}$ and the remaining\\ \hspace{1cm}$k-\left|SUR_{tar}\right|$ positions are supplemented with \\ 
\hspace{1cm}dumb points.
\STATE \hspace{0.5cm}\textbf{return}  $NUR_{tar}$
\end{algorithmic}
\label{alg2}
\end{algorithm}
\paragraph{Sort The High-Order Adjacent Roads}
In order to obtain more spatial information near the target road, multiple virtual intersections are formed into a simplified road network. According to Algorithm 3, the $d$-order upstream adjacent roads of any target road can be transformed into an ordered set with $k^d$ elements.
\begin{algorithm}[H]
\caption{Sort the $d$-order upstream adjacent roads($d>1$)}\label{alg:alg3}
\begin{algorithmic}[1]
\REQUIRE Target road $r_{tar}$; The set of 1-order upstream adjacent roads of the target road $UR_{tar}^1$;The normative number $k$ of adjacent roads of the target road;The ordered set of $(d-1)$-order upstream adjacent roads of the target road $U_{tar}^{d-1}$, specially, $U_{tar}^1\gets{NORMUP\left(r_{tar},UR_{tar}^1,k\right)}$. 
\ENSURE The ordered set of $d$-order upstream adjacent roads of the target road $U_{tar}^d$. 
\STATE \hspace{0.5cm}Initializing $U_{tar}^d\gets{\emptyset}$ and $z\gets{0}$
\STATE \hspace{0.5cm}\textbf{while} $z<\left|U_{tar}^{d-1}\right|$ \textbf{do}
\STATE \hspace{1cm}$z\gets{z+1}$
\STATE \hspace{1cm}Finding the $z$th road $r_z$ in the ordered set of\\
\hspace{1cm}$(d-1)$-order upstream adjacent roads of the target \\
\hspace{1cm}road $U_{tar}^{d-1}$, $r_z=U_{tar}^{d-1}[z]$  
\STATE \hspace{0.5cm}\textbf{if} $r_z$ is a dumb point \textbf{do}
\STATE \hspace{1cm}$U_{tar}^d\gets{U_{tar}^d}\cup{DS}$, where $DS$ is the set of $k$ \\
\hspace{1cm}dumb points
\STATE \hspace{0.5cm}\textbf{else}
\STATE \hspace{1cm}$U_{tar}^d\gets{U_{tar}^d}\cup{NORMUP\left(r_{z},UR_{z}^1,k\right)}$
\STATE \hspace{0.5cm}\textbf{return}  $U_{tar}^d$
\end{algorithmic}
\label{alg3}
\end{algorithm}
The functions of the downstream intersection and the upstream intersection are different, but the structure is that one target road corresponds to multiple adjacent roads. So the same process can be used to get the ordered set of $d$-order downstream adjacent roads of the target road $D_{tar}^d$.
\subsubsection{Convert Set to Input Matrix}
As shown in Fig. 3(a), $U_{tar}^w,\ldots,U_{tar}^1,D_{tar}^1,\ldots,D_{tar}^w$ are the ordered sets of adjacent roads of each order after the intersection reconstruction. All sets are transformed into matrices according to the rules shown in Fig. 3(b). For any $r_i\in{InputSet}$, $S_i^t=\left[speed_i^{t-h+1},\ldots,speed_i^{t}\right]$ is the historical speed sequence of the target road $r_i$ at time $t$ with a length of $h$. Obtain the historical speed sequence of all roads in the set, and then stack them in the order of the ordered set to get a historical speed matrix $M$ corresponding to the set. And $M\in{\mathbb{R}^{m\times{h}}}$, $m=k^d$, where $k$ is the normative number of the adjacent roads, $d$ is the order of the adjacent roads.

\subsection{Feature Extraction Module}
The historical data sequence of the target road only contains temporal information, while the historical data matrix of the adjacent roads contains spatialtemporal information. Therefore, the feature extraction module is divided into two sub-modules, one is the target road network, and the other is the adjacent road network.

\subsubsection{Target Road Network}
The structure of the target road network is shown in Fig. 5(a). The input data of the target road network is the historical speed sequence of the target road with a length of $h$, $S_{tar}^t=\left[speed_{tar}^{t-h+1},\ldots,speed_{tar}^t\right]$, $S_{tar}^t\in{\mathbb{R}^{1\times{h}}}$. After the sequence passes through the LSTM module, the results of each LSTM unit are spliced together. Finally, a fixed length feature vector ${trf}^t$ of the target road is obtained through the fully connected layer, and the dimension of ${trf}^t$ is $d_{hid}$.

\begin{figure}[!t]
\centering
\subfloat[]{\includegraphics[width=1.4in]{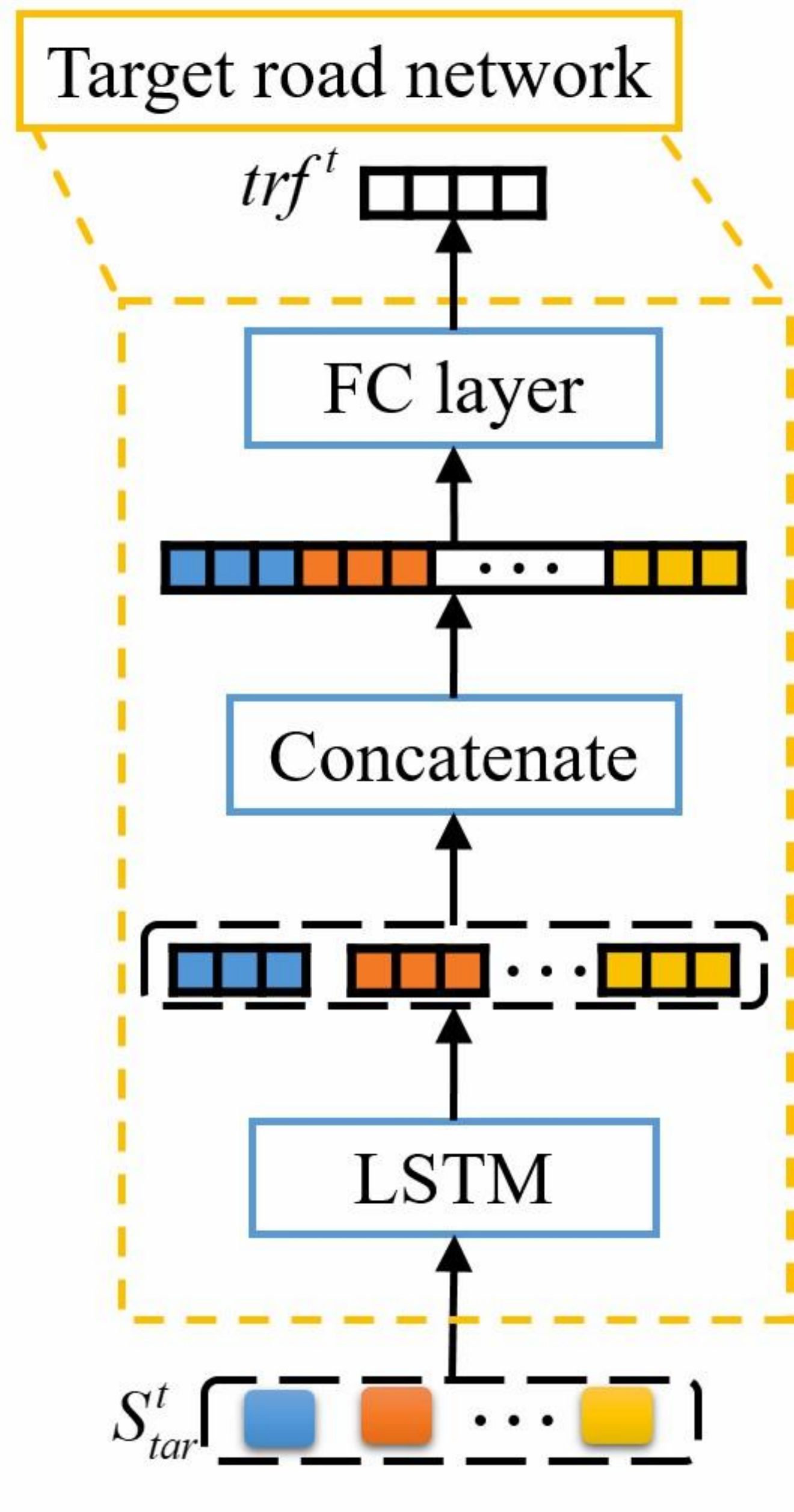}%
\label{fig5a}}
\hfil
\subfloat[]{\includegraphics[width=1.7in]{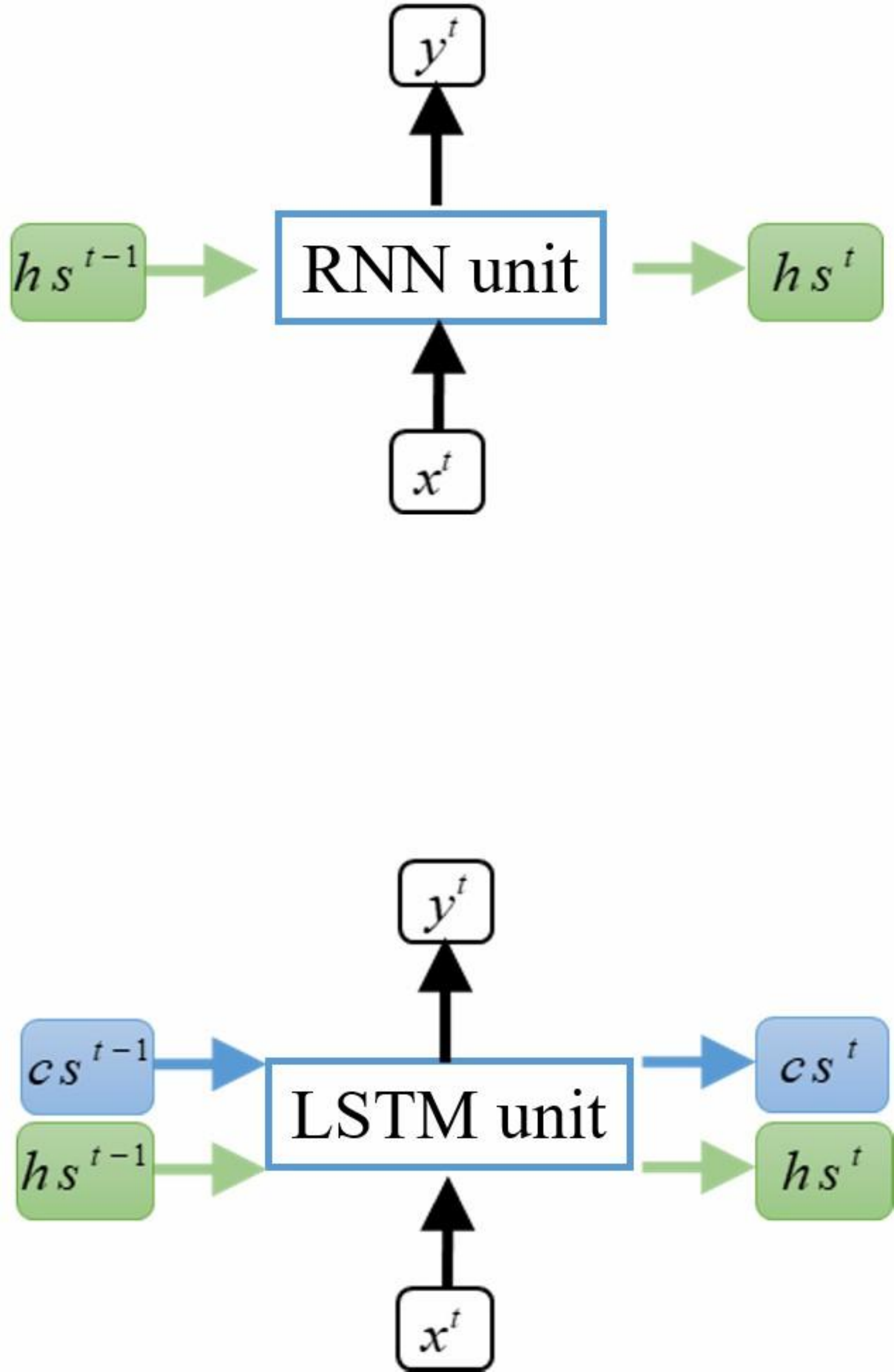}%
\label{fig5b}}
\hfil
\subfloat[]{\includegraphics[width=3in]{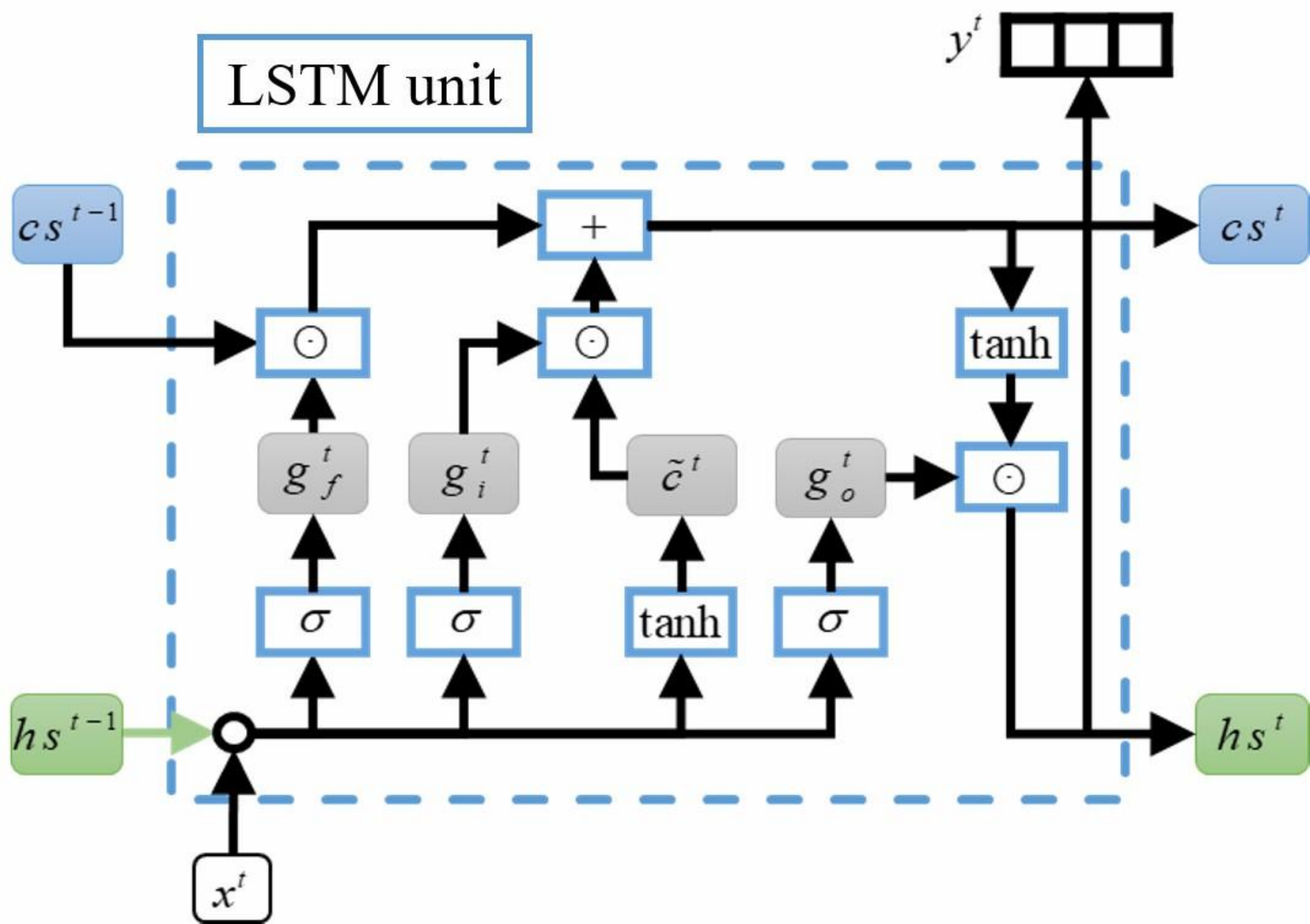}%
\label{fig5c}}
\caption{(a) The structure of the target road network. (b) The recurrent unit of RNN and LSTM. (c) The structure of LSTM unit.}
\label{fig_5}
\end{figure}

Recurrent Neural Network (RNN) as shown in Fig. 5(b) can use historical information to learn the current prediction task, which is more suitable for sequential data than other neural networks. In an ordinary RNN, there is only a simple structure, such as a tanh layer. So when the interval between the historical information and the current predicted position becomes larger, it is difficult for the RNN to learn the correlation between two steps far apart due to gradient vanishing or gradient exploding.

Whereas RNN have only one transfer state, LSTM as shown in Fig. 5(b) have two transfer states, a cell state $cs$ and a hidden state $hs$. Meanwhile, LSTM has three more gates (forget gate $g_f$, input gate $g_i$, output gate $g_o$) as shown in Fig. 5(c). The calculation formula for each gate is as follows:

\begin{equation}
\label{eq3}
g_f^t=\sigma{\left(W_f\begin{bmatrix}
x^t\\
h^{t-1}
\end{bmatrix}
+b_f\right)}
\end{equation}

\begin{equation}
\label{eq4}
g_i^t=\sigma{\left(W_i\begin{bmatrix}
x^t\\
h^{t-1}
\end{bmatrix}
+b_i\right)}
\end{equation}

\begin{equation}
\label{eq5}
g_o^t=\sigma{\left(W_o\begin{bmatrix}
x^t\\
h^{t-1}
\end{bmatrix}
+b_o\right)}
\end{equation}

\begin{equation}
\label{eq6}
\tilde{cs}^t=tanh{\left(W_c\begin{bmatrix}
x^t\\
h^{t-1}
\end{bmatrix}
+b_c\right)}
\end{equation}
where $x^t$ is the data at time $t$ in the input sequence. $W_f$, $b_f$ are the weight parameters and bias parameters of the forget gate. $W_i$, $b_i$ are the weight parameters and bias parameters of the input gate. $W_o$, $b_o$ are the weight parameters and bias parameters of the output gate. $\sigma$ denotes the logistic sigmoid function. $\tilde{cs}^t\in{[-1,1]}$ is the updated value of the cell state. $W_c$, $b_c$ are the weight parameters and bias parameters of the update neural network layers. $g_f^t$, $g_i^t$ and $g_o^t$ are respectively the value of the forget gate, the input gate and the output gate at time $t$, $g_f^t,g_i^t,g_o^t\in{[0,1]}$. The forget gate $g_f^t$ is used to control the cell state of the previous state $cs^{t-1}$. The input gate $g_i^t$ selects the input $\tilde{cs}^t$. The output gate $g_o^t$ is used to control the cell state $cs^t$ to output at the end. $cs^t$ and $hs^t$ are obtained as follows:
\begin{equation}
\label{eq7}
cs^t=g_f^t\odot{cs^{t-1}}+g_i^t\odot{\tilde{cs}^t}
\end{equation}

\begin{equation}
\label{eq8}
hs^t=g_o^t\odot{tanh(cs^t)}
\end{equation}
where $\odot$ is Hadamard product, which is the multiplication of the corresponding elements in the matrix. It can be seen from the formula that $cs$ generally changes slowly, and $hs$ is very different from the previous moment. Based on multiple gating mechanisms and two transit states, LSTMs are more capable of capturing long-term dependencies.
\subsubsection{Adjacent Road Network}
As shown in Fig. 6(a), the structure of the adjacent road network is divided into three parts: the convolutional layer to extract intersection features, the Temporal-LSTM to extract temporal features, and the Spatial-LSTM to extract spatial features. The input matrix of the $d$-order adjacent road is processed by the convolution layer and Temporal-LSTM to obtain the temporal feature $tf_d^t$ of the $d$-order adjacent road at time $t$. The sequence is formed by the temporal features of adjacent roads of each order according to the sequence of vehicle flow. The spatial feature of adjacent roads of each order is obtained through Spatial-LSTM. The spatial feature of the $d$-order adjacent road at time $t$ is recorded as $sf_d^t$.

\begin{figure}[!t]
\centering
\subfloat[]{\includegraphics[width=1.7in]{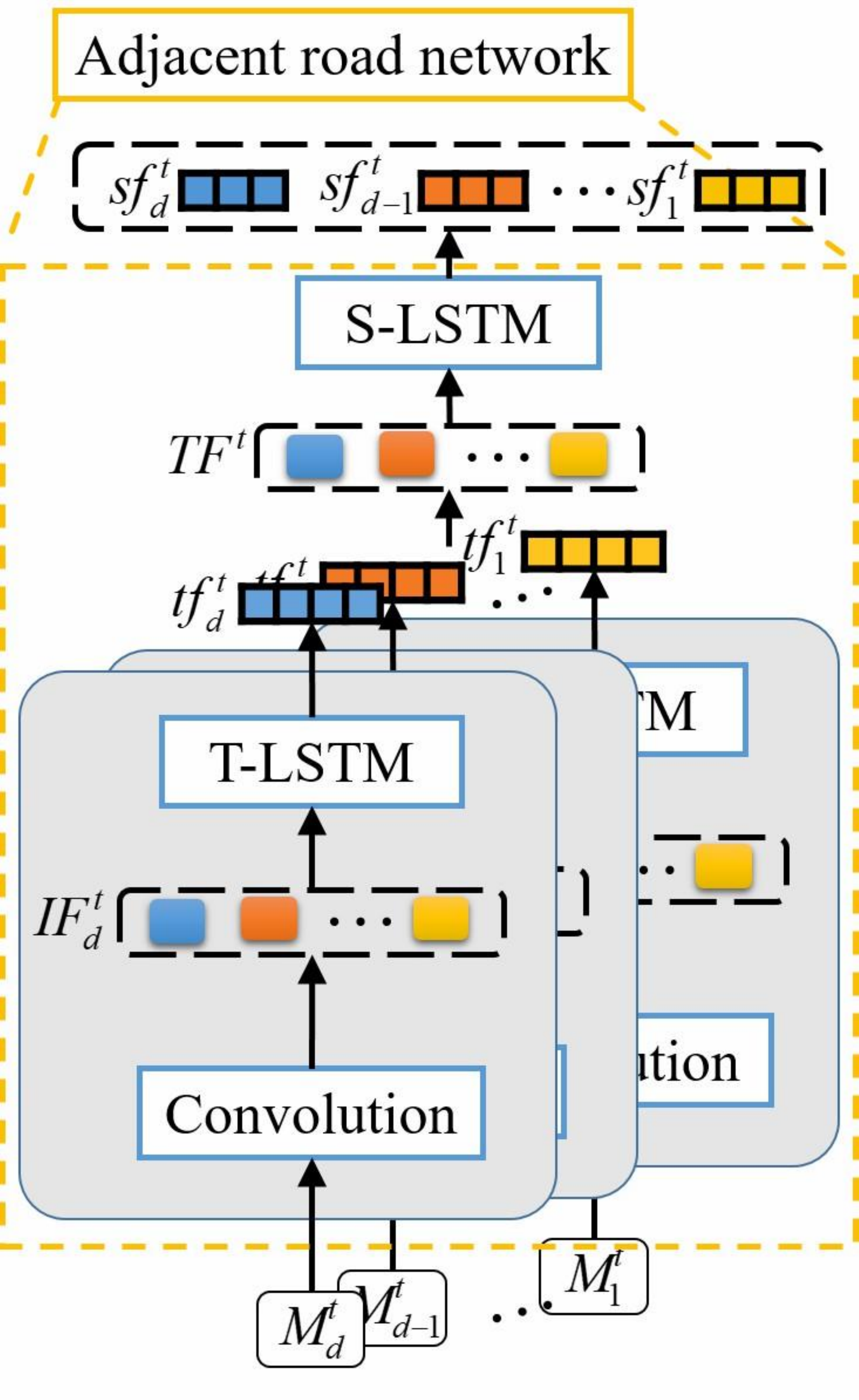}%
\label{fig6a}}
\hfil
\subfloat[]{\includegraphics[width=1.7in]{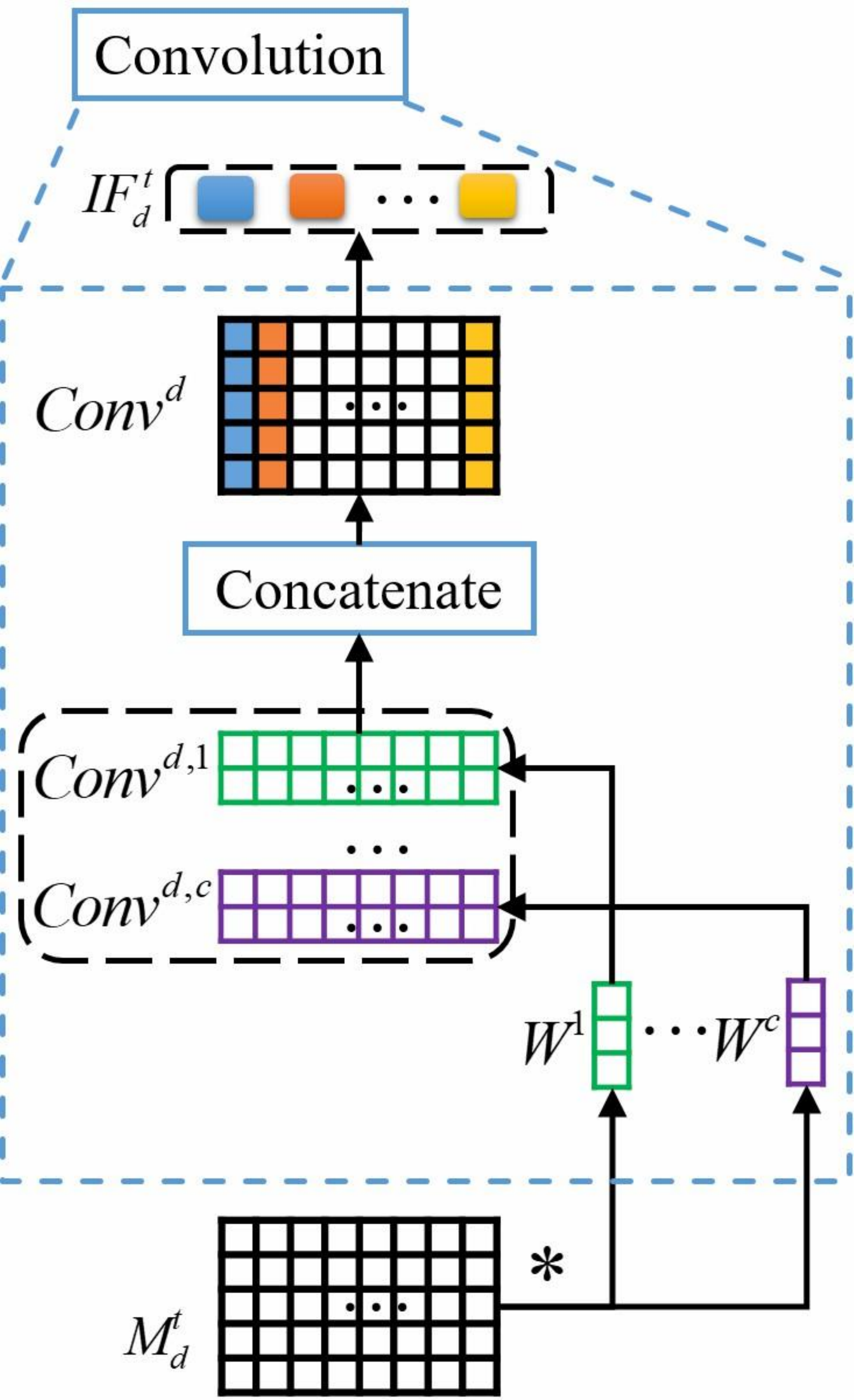}%
\label{fig6b}}
\caption{(a) The structure of the adjacent road network. (b) The specific operation of the convolutional layer.}
\label{fig_6}
\end{figure}

\paragraph{the convolutional layer}
Known from the data generation module, the stacking order of the historical speed matrix from top to bottom is consistent with the order of the ordered set. Therefore, every $k$ row in the matrix is the historical speed data of an intersection, and the adjacent roads with the same relative position have similar roles in the intersection. Convolutional neural network extracts data features by sliding the convolution kernel on the data. The convolution kernel locally connects neurons and input data, and the weights of the convolution kernels are shared. These properties make convolutional neural networks particularly suitable for extracting the features of a single intersection at each moment in the input matrix. The convolution process can be expressed according to the structure of the intersection by the following formulas:
\begin{equation}
\label{eq9}
Conv_{U,d}^c=\sigma{\left(W_U^c\ast{UM_d^t}+b_U^c\right)}
\end{equation}

\begin{equation}
\label{eq10}
Conv_{D,d}^c=\sigma{\left(W_D^c\ast{DM_d^t}+b_D^c\right)}
\end{equation}
where $UM_d^t$ is the historical speed matrix of the $d$-order upstream adjacent road at time $t$. $W_U^c$, $b_U^c$ are the weight parameters and bias parameters of the $c$-th convolution channel of the upstream intersection. $Conv_{U,d}^c$ is the output of the historical speed matrix of the $d$-order upstream adjacent road after passing through the $c$-th convolution channel. The convolution formula of the downstream adjacent road network is the same.

The specific operation of the convolutional layer is shown in Fig. 6(b). Taking the upstream road as an example, the processing method of the downstream road is the same. In order to make the convolution kernel exactly correspond to the data of one intersection at one moment, take $W_U^c\in{\mathbb{R}^{k\times{1}}}$. The stride of the convolution kernel is set to $[k,1]$. It means the distance that the convolution kernel slides on the data after each convolution operation. It can be calculated $Conv_{U,d}^c\in{\mathbb{R}^{(m\div{k})\times{h}}}$, $m=k^d$. At the same time, because the convolution kernel size and stride do not stretch the matrix in the time dimension, the length is still $h$. The convolutional layer in this paper obtains different features of each intersection structure through multiple convolutional channels. Then the multi-channel intersection features are stacked together to construct the intersection feature matrix. Treat the column data of the feature matrix as the feature vector $uif_d^t$ of the corresponding time, a sequence of intersection features $UIF_d^t=\left[uif_d^{t-h+1},\ldots,uif_d^t\right]$ can be obtained.

Since the features of the intersection structure are learned, the historical speed matrix of the upstream adjacent roads of different orders can extract the intersection features through the convolution layer of the same structure and the same parameters. It should be noted that the input matrices of different orders have different sizes, so $uif$ of different orders have different dimensions.

\paragraph{LSTM Layer}
The core of both Temporal-LSTM and Spatial-LSTM is the basic LSTM structures as described in the target toad network. The main difference between temporal LSTM and spatial LSTM is that the input data are different, so different features can be captured. At the same time, the two methods are slightly different in other parts except LSTM.

The input of the Temporal-LSTM is the sequence of intersection features obtained by the convolutional layer $IF_d^t=\left[if_d^{t-h+1},\ldots,if_d^t\right]$. The sequence length is $h$, and each moment has $c\times{k^{d-1}}$ elements. The structure of Temporal-LSTM is the same as the structure of the target road network. The outputs of all LSTM units are spliced together, and the fully connected layer outputs a temporal feature vector $tf_d^t$ of the $d$-order adjacent roads at time $t$ with a fixed length.

The speed of the road is closely related to the flow of vehicles in the road network, and the flow direction of the vehicles is unique. Therefore, the temporal feature vector $tf_d^t$ of the adjacent roads can be formed into a spatial sequence in the order from high order to low order, $TF^t=\left[tf_d^t,\ldots,tf_1^t\right]$. The spatial sequence is processed by LSTM, and the output of each LSTM unit is taken as the spatial feature vector $sf_d^t$ of adjacent roads of the corresponding order. The dimensions of all spatial feature vectors are the same as the target road output vector $trf^t$, which is $d_{hid}$.

\paragraph{Feature Fusion Module}
\begin{figure}[!t]
\centering
\includegraphics[width=3in]{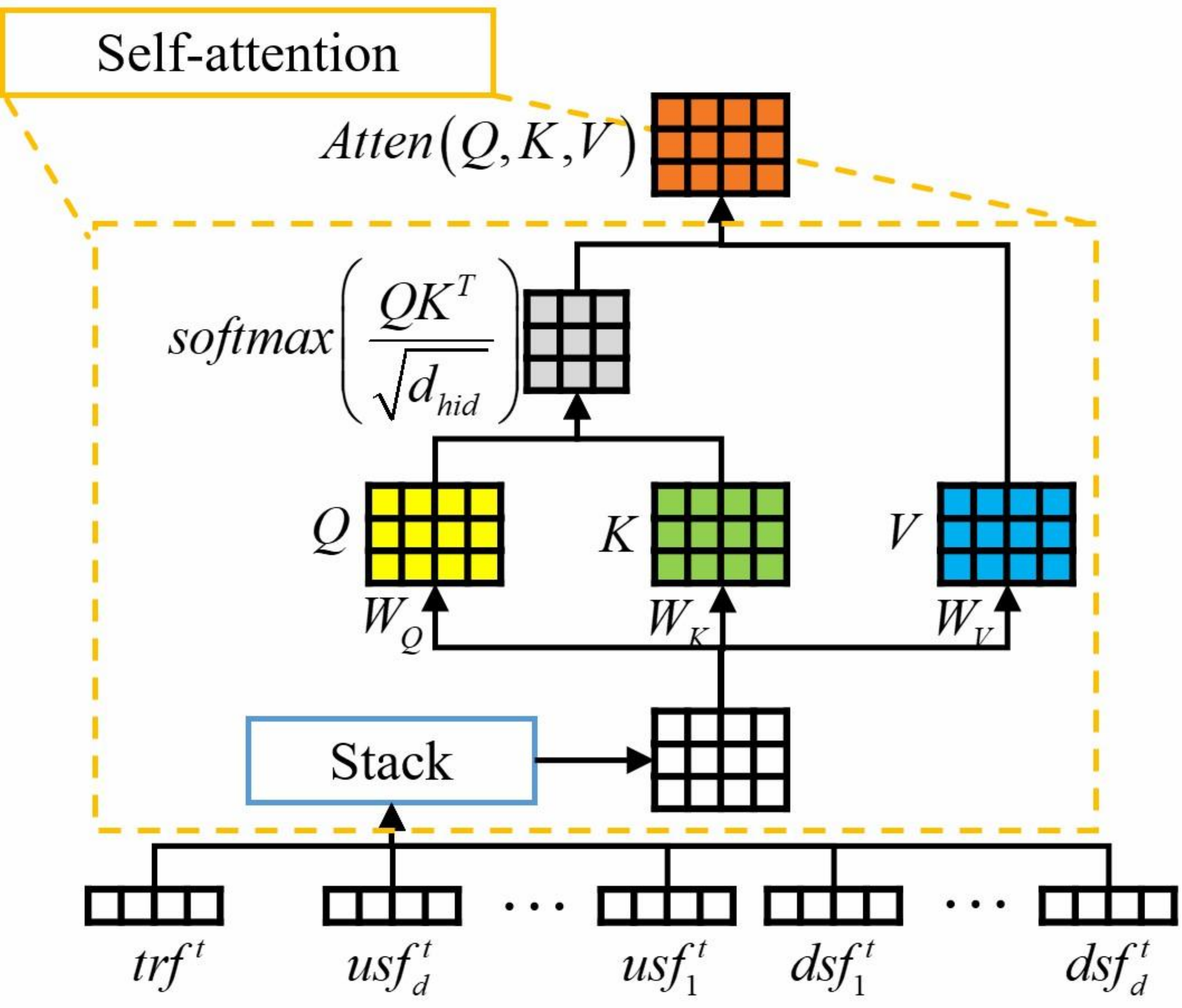}
\caption{The calculation process of the self-attention mechanism.}
\label{fig7}
\end{figure}
In order to reflect the influence of the adjacent roads of each order on the target road, this paper combines the self-attention mechanism to quantify the relationship among the roads of each order. Through three different fully connected layers, the spatial feature vector $sf_d^t$ can be mapped into three vectors respectively called Query $Q_d$, Key $K_d$, Value $V_d$. The value of the dot product of $Q_i$ and $K_j$ indicates the correlation between $sf_i^t$ and $sf_j^t$. Next, divide the value by the square root of the vector dimension $d_{hid}$ and use the softmax function to get the attention weight. Finally, the attention weight is multiplied by $V_d$. For computational efficiency, the vectors can be stacked as a matrix for calculation as shown in Fig. 7, $Q\in{\mathbb{R}^{(2d+1)\times{d_Q}}}$, $K\in{\mathbb{R}^{(2d+1)\times{d_K}}}$, $V\in{\mathbb{R}^{(2d+1)\times{d_V}}}$. The formula is as follows:
\begin{equation}
\label{eq11}
Atten(Q,K,V)=softmax\left(\frac{QK^T}{\sqrt{d_{hid}}}\right)V
\end{equation}
where $d_{hid}$ is the vector dimension for calculating attention weights, in this paper, $d_Q=d_K=d_V=d_{hid}$.

Finally, the obtained result is flattened and the final output is obtained through the fully connected layer.
\section{Experiment}
\subsection{Preparation}
\subsubsection{Source Dataset}
This paper evaluates the model with the Caltrans Performance Measurement System (PeMS), a real-world traffic dataset. A total of 39,000 sensors are deployed on highway systems in all major urban areas of California, as shown in Fig. 8(a). A region of California was used, as shown in Fig. 8(b). And the data from January 1, 2020 to January 31, 2020 recorded every hour are used as the source data.
\subsubsection{Data Preprocessing}
The opposite lanes are regarded as two different roads, and there are 234 roads in the selected area. Match each road with the sensors installed on the road, and take the average speed of all sensors corresponding to a road to represent the speed of the road. The interval time is taken as 1 hour, that is to say, each road contains 744 historical speed records.
\begin{figure}[!t]
\centering
\subfloat[]{\includegraphics[width=1.5in]{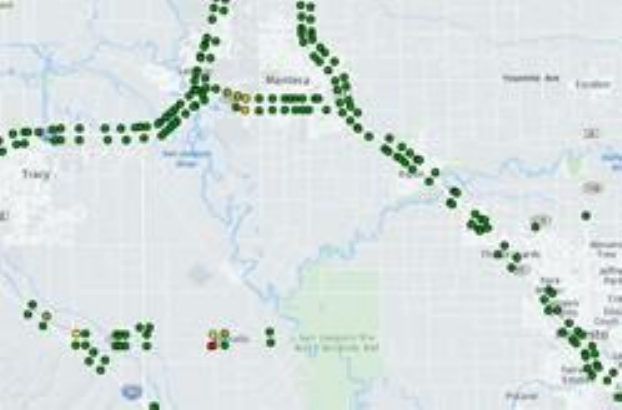}%
\label{fig8a}}
\hfil
\subfloat[]{\includegraphics[width=1.5in]{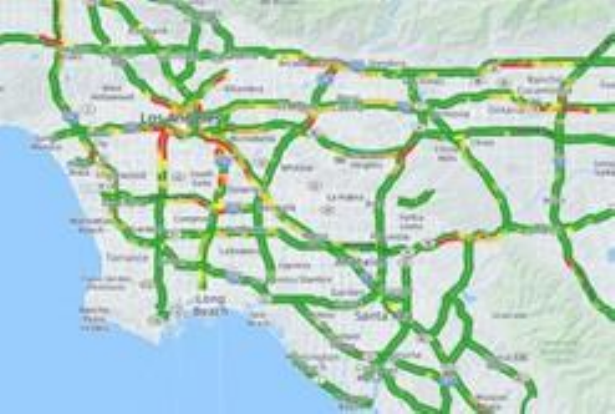}%
\label{fig8b}}
\caption{(a) Scenarios for Sensor Deployment. (b) The data area selected in this paper.}
\label{fig_8}
\end{figure}
\subsubsection{Evaluation}
The effect evaluation method in this paper is Root Mean Square Percentage Error (RMSPE) and Mean Absolute Percentage Error (MAPE), and the formula is as follows:
\begin{equation}
\label{eq12}
RMSPE=\sqrt{\frac{1}{m}\sum_{i=1}^m\left(\frac{y_i-\hat{y}_i}{y_i}\right)^2}\times{100\%}
\end{equation}
\begin{equation}
\label{eq13}
MAPE=\frac{1}{m}\sum_{i=1}^m\left|\frac{y_i-\hat{y}_i}{y_i}\right|\times{100\%}
\end{equation}
where $y_i$ is the actual value and $hat{y}_i$ is the predicted value.
\subsection{Parameter}
\subsubsection{Model Parameter}
In the data generation module, the parameter that need to be defined is the normative number $k$. In this paper, $k=3$, which means that an intersection contains one target road and three adjacent roads.

The parameters of the training process are shown in Table 1. The main components of the target road network are one LSTM layer and one fully connected layer, and the parameters are shown in Table 2. The main components of the adjacent road network are a convolutional layer, one LSTM layer and one fully connected layer of Temporal-LSTM, and one LSTM layer of Spatial-LSTM. The parameters are shown in Table 3. 

\begin{table}[!t]
\caption{Parameters of the training process}
\centering
\begin{tabular}{|c||c|}
\hline
Training Parameter & \\
\hline
Train road & 10244\\
The proportion of training set & 60\%\\
The proportion of validation set & 20\%\\
The proportion of test set & 20\%\\
Learning rate & 0.001\\
Dropout & 0\\
Batch size & 16\\
Early stop epoch & 500\\
Normalization & minmax\\
\hline
\end{tabular}
\end{table}
\begin{table}[!t]
\caption{Parameters of the target road network}
\centering
\begin{tabular}{|c||c|}
\hline
Parameters of the target road network & \\
\hline
The number of layers of LSTM & 2\\
The number of features in the hidden layer & 256\\
The output dimension of the fully connected layer & 256\\
\hline
\end{tabular}
\end{table}
\begin{table}[!t]
\caption{Parameters of the adjacent road network}
\centering
\begin{tabular}{|c||c|}
\hline
Parameters of the adjacent road network & \\
\hline
The size of convolution kernel & [3,1]\\
The stride of convolution kernel & [3,1]\\
The number of convolution channels & 6\\
Padding & [0,0]\\
The number of layers of T-LSTM & 2\\
The number of features in the hidden layer of T-LSTM & 512\\
The output dimension of the fully connected layer of T-LSTM & 32\\
The number of layers of S-LSTM & 2\\
The number of features in the hidden layer of S-LSTM & 256\\
\hline
\end{tabular}
\end{table}
\subsubsection{Input Parameter}
As mentioned in Section \uppercase\expandafter{\romannumeral3}, there are two important input parameters for the input data of the model. One is the space width $w$, and the other is The length of historical sequence $h$. According to the definition, it can be seen that the larger $h$ and $w$ are, the more spatiotemporal information is input into the model. In general, more spatiotemporal features can be captured from more spatiotemporal information, and predictions are more accurate. However, the calculation cost of the model increases accordingly, as shown in Table 4.
\begin{table}[!t]
\caption{Floating point operations (FLOPs) of the model under different input parameters}
\centering
\begin{tabular}{|c||c c c c c|}
\hline
\diagbox{w}{h} & 2 & 3 & 4 & 5 & 6\\
\hline
1 & 257.60M & 373.06M & 488.52M & 603.98M & 719.43M\\
2 & 489.56M & 707.68M & 925.80M & 1.14G & 1.36G\\
3 & 727.46M & 1.05G & 1.37G & 1.70G & 2.02G\\
\hline
\end{tabular}
\end{table}
The goal is to use as small $w$ and $h$ as possible to obtain a better prediction effect. This paper sets $h\in{[2,3,4,5,6]}$ and $w\in{[1,2,3]}$. The result for $p\in{[1,2,3,4,5]}$ is shown in Fig. 9. When $w$ remains constant and $h$ increases, the result is shown in Figure 9(a-c). With the increase of $h$, the overall speed prediction effect for the future time is constantly getting better. Especially for the input with larger $w$, increasing $h$ has a more obvious effect on the prediction result. In the same way, when $h$ continues to increase, the prediction effect is not significantly improved and the amount of calculation is increased, so the experiments after this paper set $h=6$.
\begin{figure}[!t]
\centering
\subfloat[]{\includegraphics[width=1.7in]{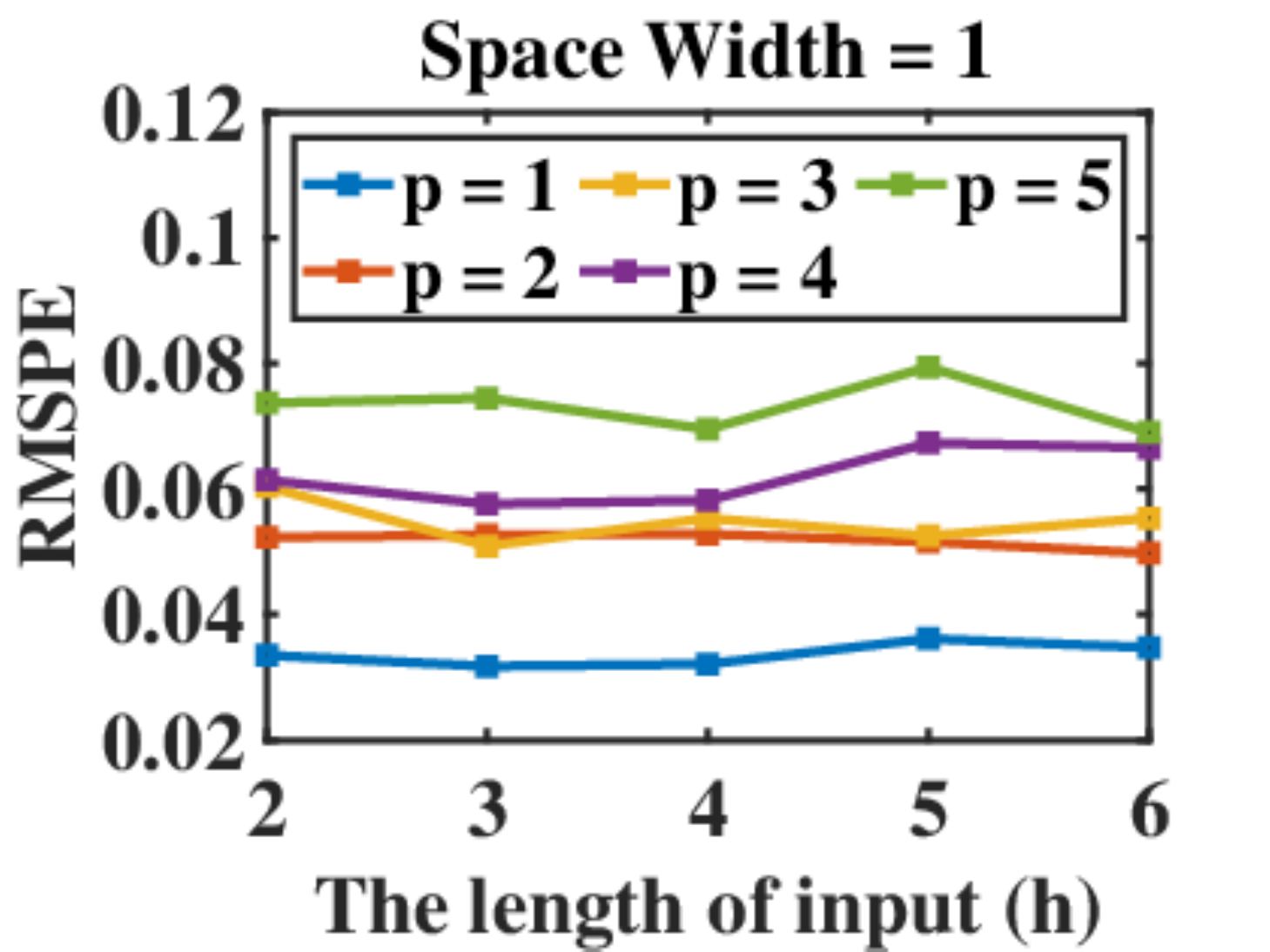}%
\label{fig9a}}
\hfil
\subfloat[]{\includegraphics[width=1.7in]{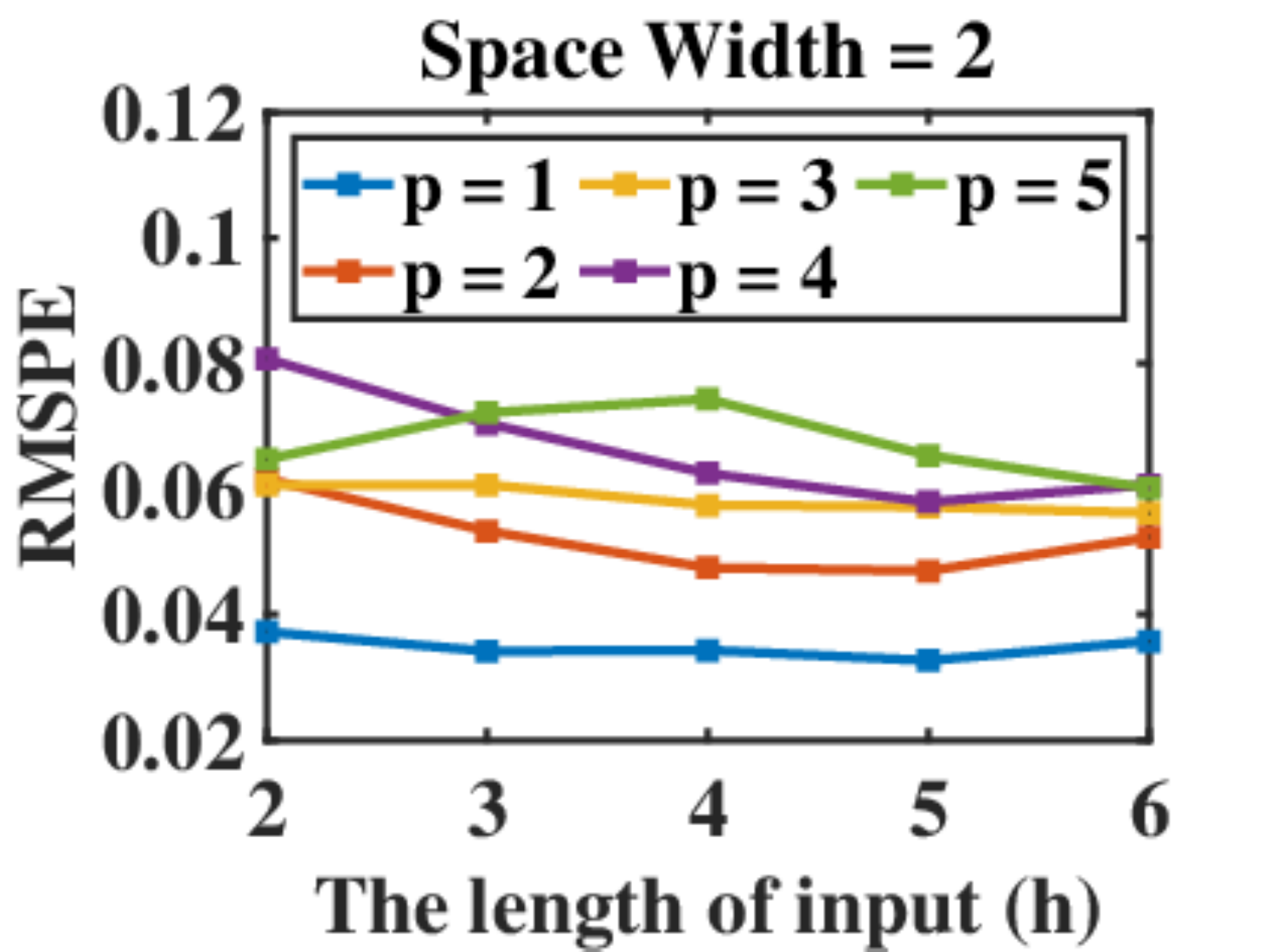}%
\label{fig9b}}
\hfil
\subfloat[]{\includegraphics[width=1.7in]{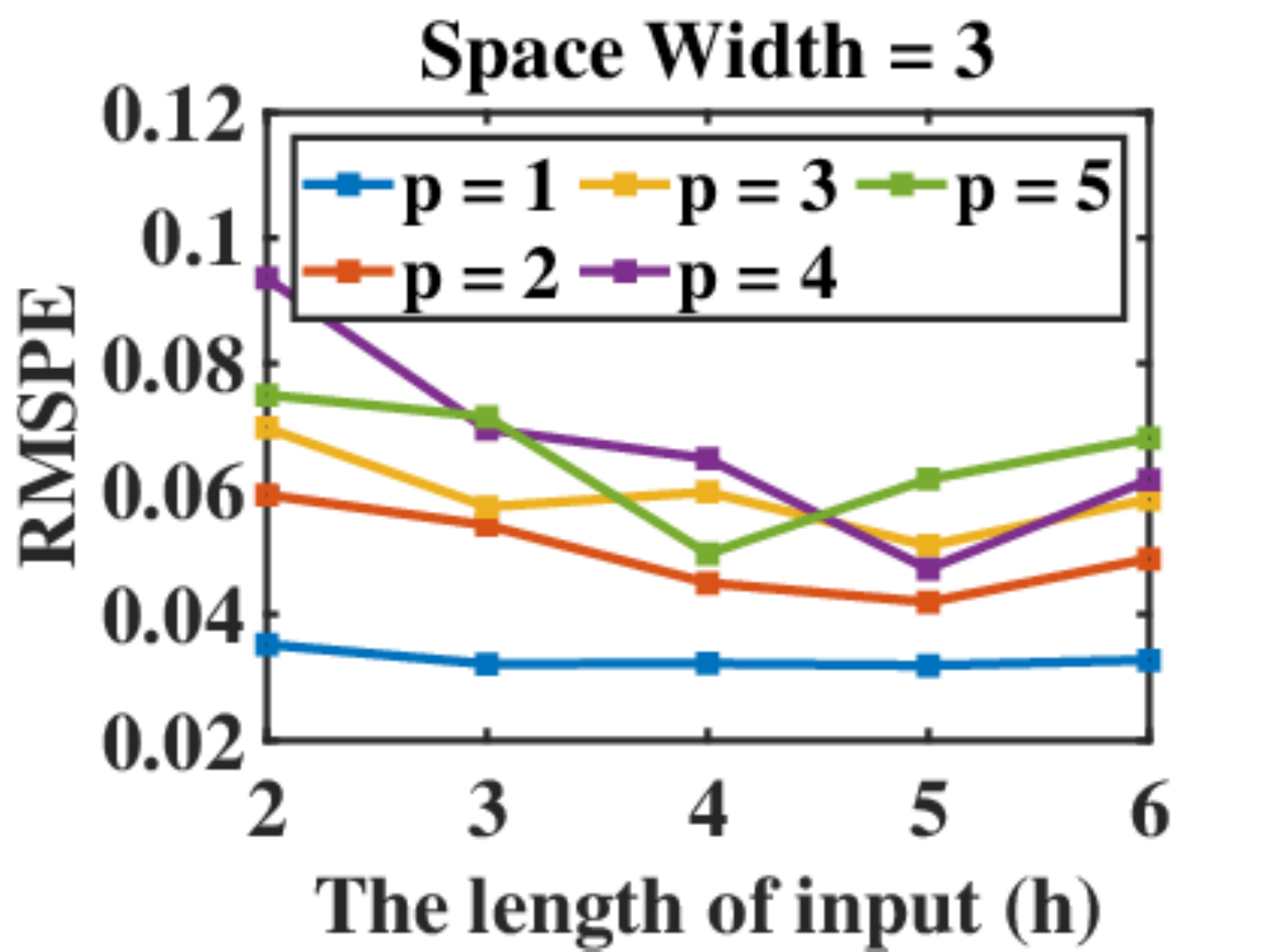}%
\label{fig9c}}
\hfil
\subfloat[]{\includegraphics[width=1.7in]{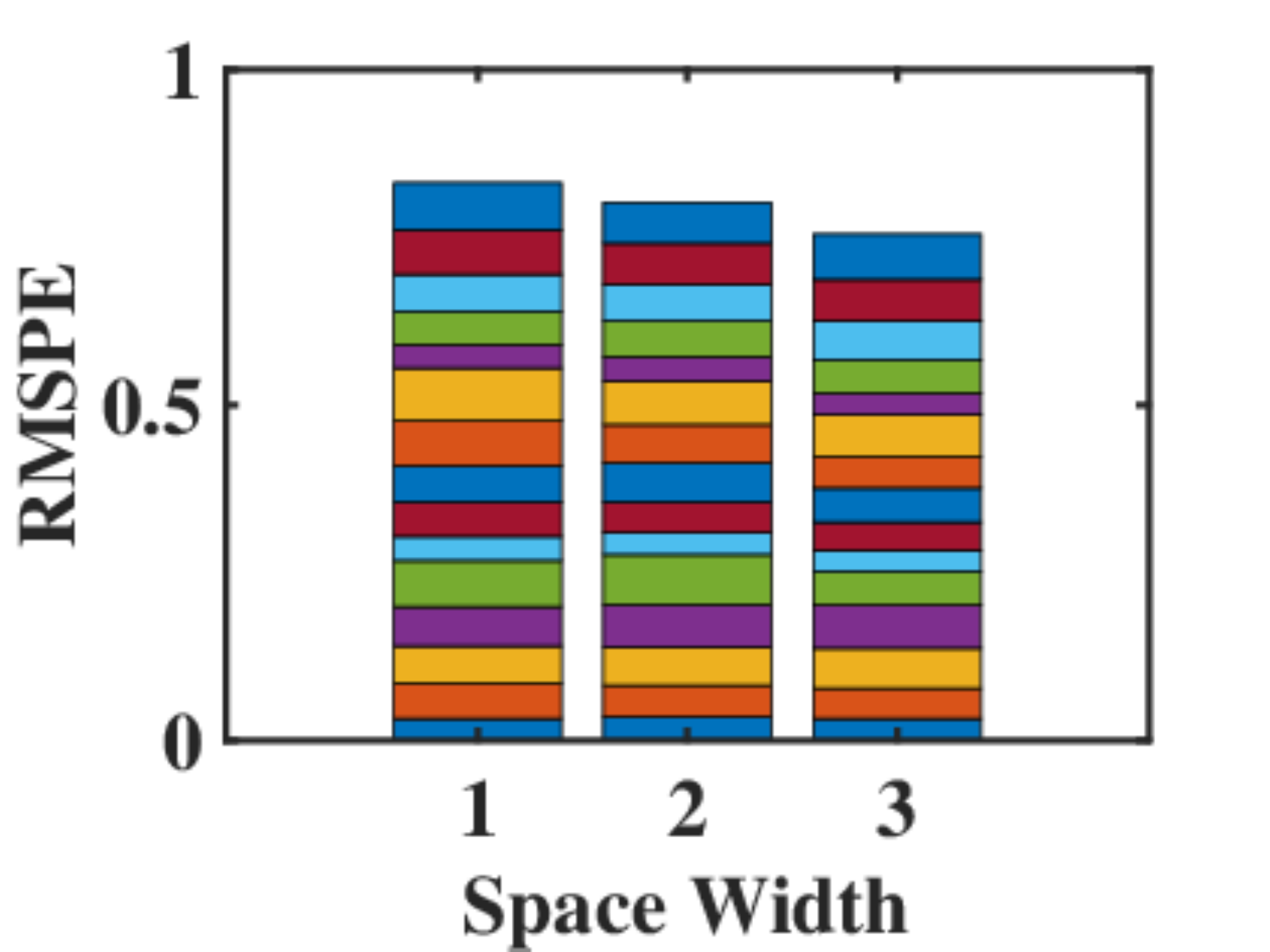}%
\label{fig9d}}
\caption{(a) Scenarios for Sensor Deployment. (b) The data area selected in this paper.}
\label{fig_9}
\end{figure}
Conversely, when $h$ is small, increasing $w$ will lead to worsening of the effect. This is because in a short time, the traffic flow of the adjacent road far from the target road has less influence on the target road. In this paper, all 15 prediction results with $h>3$ are stacked as shown in Fig. 9(d). As $h$ increases, increasing $w$ improves the prediction effect significantly. Due to hardware limitations, $w$ of all subsequent experiments in this paper is set to 3.
\subsection{Comparison with Baselines}
For sake of evaluating the prediction performance of irnet, five benchmark methods are used. Among them, ARIMA can only input the historical speed sequence of the target road. The input of CNN, LSTM, SAEs, ResMLP can be the input matrix of the target road and adjacent roads, but such an input method cannot be transferred because of the different number of roads. Finally, the input data of the baseline methods are the historical speed sequence of the target road. All baseline methods are optimized to a best performance.

ARIMA: It is a method based on time series analysis and is a classic statistical model for predicting traffic flow. The number of autoregressive terms, the number of nonseasonal differences needed for stationarity and the number of lagged forecast errors in the prediction equation are set to 5, 2 and 0. That is, using the ARIMA (5, 2, 0) forecasting equation.

CNN: It refers to a CNN-based traffic flow prediction model. This paper uses a single-layer CNN, the size of the convolution kernel is [3, 1], the stride of the convolution kernel is [1, 1], with no padding, and the number of channels is 16. Finally, a multi-task linear regression model is trained on the top layer.

LSTM: This paper uses a three-layer LSTM layer with 512 hidden units to process the time series. Flatten the output of the LSTM unit at all moments, and finally trains a multi-task linear regression model on the top layer.

SAEs: Stacked autoencoders are configured with three layers of [64, 256, 512] hidden units for pre-training. Finally train a multi-task linear regression model on the top layer.

ResMLP: As one of the most advanced deep learning algorithms\cite{44}, this paper simply applies it to traffic speed prediction. The data at each time is regarded as a picture with only one pixel, and the length of the time series is regarded as the number of channels of the picture. The hidden layer unit is set to 256 and the network depth is set to 12.

The results are shown in Table 5. The prediction effect of IRNet is better than that of the baseline method, and it has a good performance in the speed prediction of short prediction horizon. At the same time, LSTM, as a deep learning algorithm for processing time series, has shown better performance in speed prediction tasks with long prediction horizon.
\begin{table}[!t]
\caption{Prediction of the speed of target road}
\centering
\begin{tabular}{|c||c c c c c|}
\hline
\multicolumn{6}{|c|}{h=6   w=3}\\
\hline
p & 1 & 2 & 3 & 4 & 5\\
\hline
\multicolumn{6}{|c|}{RMSPE}\\
\hline
ARIMA & 4.563\% & 8.652\% & 12.200\% & 15.261\% & 18.011\%\\
SAEs & 6.647\% & 7.982\% & 7.637\% & 8.742\% & 9.389\%\\
CNN & 4.876\% & 6.342\% & 7.625\% & 7.982\% & 8.571\%\\
LSTM & 4.022\% & 6.227\% & 6.401\% & 6.226\% & \textbf{6.235\%}\\
ResMLP & 3.555\% & 4.961\% & 6.056\% & 6.809\% & 6.729\%\\
IRNet & \textbf{3.275\%} & \textbf{4.887\%} & \textbf{5.835\%} & \textbf{6.142\%} & 6.809\%\\
\hline
\multicolumn{6}{|c|}{MAPE}\\
\hline
ARIMA & 3.063\% & 5.995\% & 8.881\% & 11.580\% & 14.080\%\\
SAEs & 5.141\% & 6.390\% & 6.440\% & 7.073\% & 7.243\%\\
CNN & 2.943\% & 4.542\% & 5.969\% & 6.380\% & 6.661\%\\
LSTM & 3.307\% & 5.026\% & 4.483\% & 5.172\% & \textbf{4.664\%}\\
ResMLP & 2.493\% & \textbf{3.700\%} & 4.651\% & 5.150\% & 4.904\%\\
IRNet & \textbf{2.371\%} & 3.733\% & \textbf{3.488\%} & \textbf{4.548\%} & 5.229\%\\
\hline
\end{tabular}
\end{table}
It can be seen in Fig. 10 (a) that the points above y=x of IRNet are offset farther, but the offset points are less than other methods. And in Fig. 10 (b), the error of IRNet is slightly smaller than that of other methods in most of the time between 0-80H, but occasionally a larger error occurs, such as near 60H and after 80H.
\begin{figure}[!t]
\centering
\subfloat[]{\includegraphics[width=3.5in]{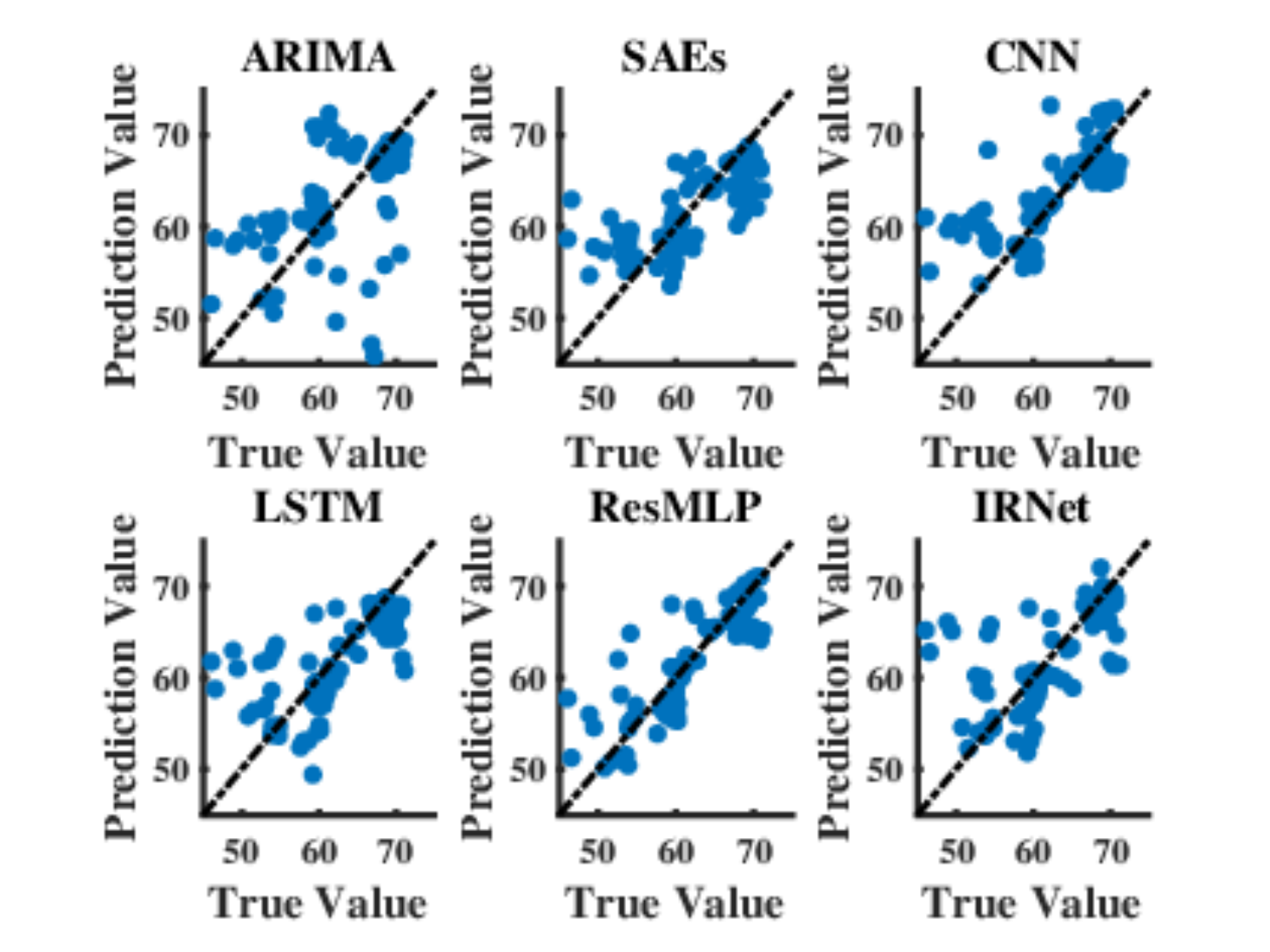}%
\label{fig10a}}
\hfil
\subfloat[]{\includegraphics[width=3.5in]{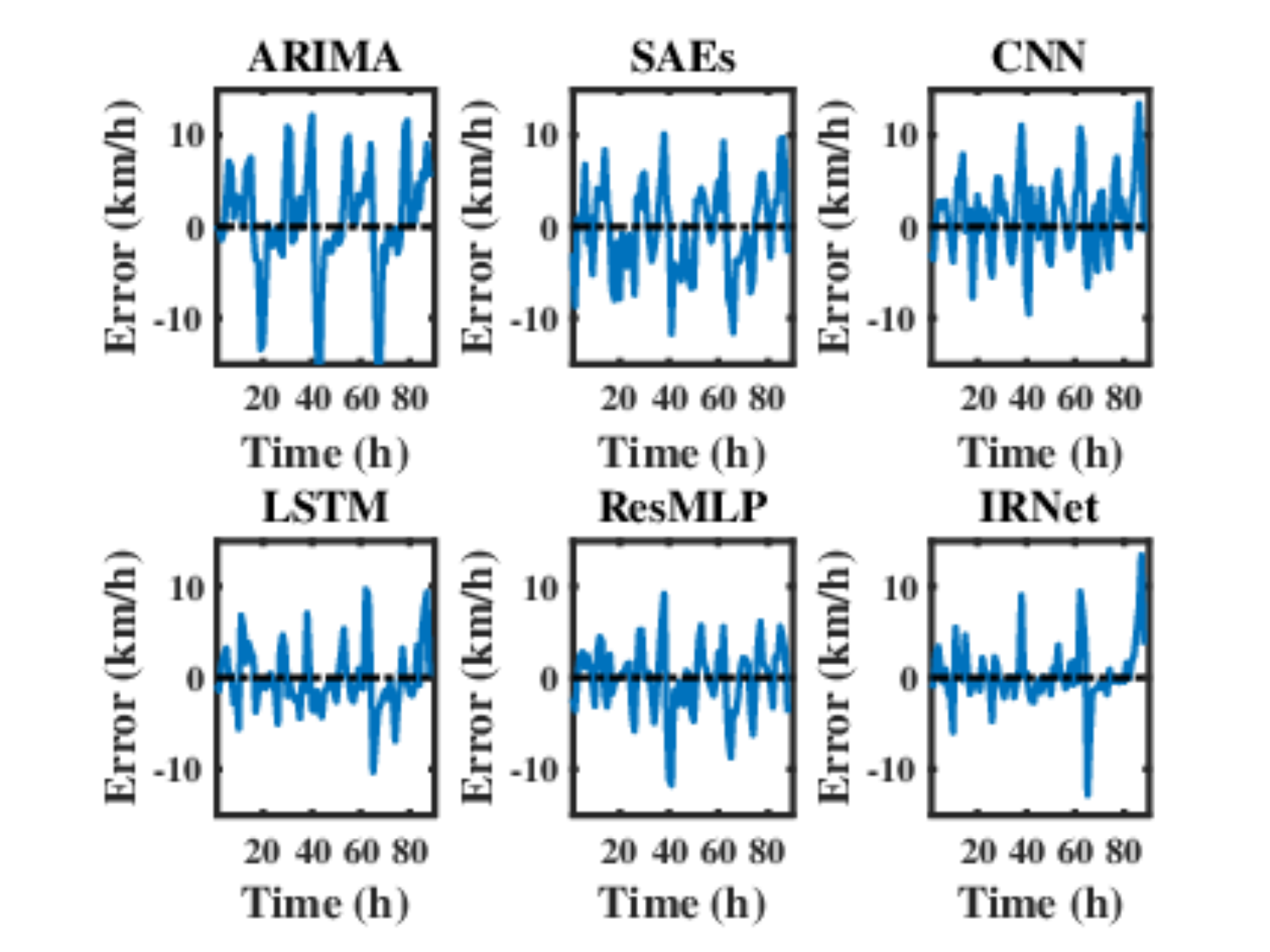}%
\label{fig10b}}
\caption{(a) The scatter plot of true values and predicted values for different methods on the test set. (b) The error plot of true values and predicted values for different methods on the test set.}
\label{fig_10}
\end{figure}
\subsection{Transfer Learning}
Transfer learning is a popular method in the field of deep learning. It can speed up the training process of the model and can effectively reduce the amount of target data required. However, the limitation of transfer learning is that it requires a large amount of raw data to train a pre-trained model suitable for the source task, learning as much as possible about the raw data. Then the new model for the target task is established and trained by transferring knowledge. Therefore, the source task and the target task must be highly correlated, and general knowledge needs to be learned from the source task. The speed prediction of different roads is a highly related task.

The baseline methods only use the historical data of target road, while the method in this paper reconstructs the road structure. Therefore, the target road does not affect the structure of the model. Invariant model structure is a precondition for model transfer. 
\begin{figure}[!t]
\centering
\includegraphics[width=3.5in]{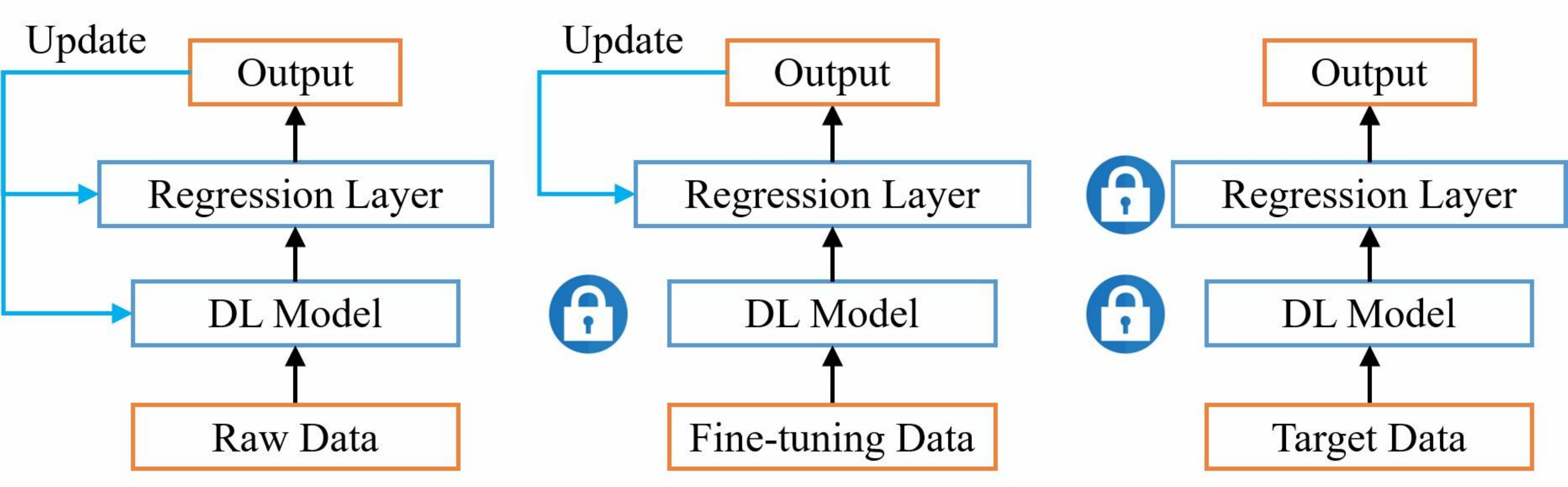}%
\label{fig11}
\caption{The overall strategy of transfer learning.}
\label{fig_11}
\end{figure}

In transfer learning, the parameters of the whole model are updated by training on the raw dataset to obtain a pre-trained model. Then, all of the model parameters are fixed. The model is fine-tuned with very little target data, and only the parameters of the final regression layer are updated. Finally, the target dataset is used to verify the prediction effect of the fine-tuned model. The overall strategy is shown in Fig. 11.

For the various baseline methods, the data of the road NO.10244 is used for training. The trained model is saved as a pre-trained model. The road NO.10384 is selected as the unknown road. And the first ten moments of the road NO.10384 are used as the fine-tuning dataset. The results are shown in Table 6. After fine-tuning, IRNet has a significant improvement over other baseline methods. This is because that baseline methods can only transfer temporal features. However, IRNet can transfer the spatiotemporal features at the same time after reconstruction of the intersection, which greatly improves the transfer ability of the model.
 
\begin{table}[!t]
\caption{Prediction of the speed of the unknown road without any processing}
\centering
\begin{tabular}{|c||c c c c c|}
\hline
\multicolumn{6}{|c|}{h=6   w=3}\\
\hline
p & 1 & 2 & 3 & 4 & 5\\
\hline
\multicolumn{6}{|c|}{RMSPE}\\
\hline
ARIMA & 3.760\% & 7.101\% & 8.794\% & 9.907\% & 11.473\%\\
SAEs & 5.595\% & 6.371\% & 6.346\% & 6.535\% & 6.551\%\\
CNN & 14.911\% & 14.591\% & 17.031\% & 16.948\% & 13.315\%\\
LSTM & 4.784\% & 5.481\% & 7.374\% & 6.009\% & 6.798\%\\
ResMLP & 4.658\% & 6.304\% & 6.534\% & 7.317\% & 7.548\%\\
IRNet & \textbf{2.701\%} & \textbf{5.269\%} & \textbf{4.837\%} & \textbf{5.956\%} & \textbf{5.568\%}\\
\hline
\multicolumn{6}{|c|}{MAPE}\\
\hline
ARIMA & 2.192\% & 4.144\% & 5.682\% & 6.873\% & 8.086\%\\
SAEs & 4.045\% & 4.568\% & 4.683\% & 4.854\% & 4.864\%\\
CNN & 8.505\% & 9.212\% & 10.178\% & 10.409\% & 9.023\%\\
LSTM & 3.798\% & 4.290\% & 5.589\% & 4.930\% & 4.878\%\\
ResMLP & 2.873\% & \textbf{4.054\%} & 4.543\% & 5.380\% & 5.722\%\\
IRNet & \textbf{2.108\%} & 4.379\% & \textbf{3.654\%} & \textbf{4.890\%} & \textbf{4.524\%}\\
\hline
\end{tabular}
\end{table}
In Fig. 12(a), the predicted values of IRNet from 55km/h to 65km/h are very close to the y=x line, but around 70km/h, the predicted values of IRNet are generally low. In Fig. 12(b), it can also be seen that the overall error values of IRNet are more evenly distributed in the positive and negative regions compared to ResMLP and LSTM.
\begin{figure}[!t]
\centering
\subfloat[]{\includegraphics[width=3.5in]{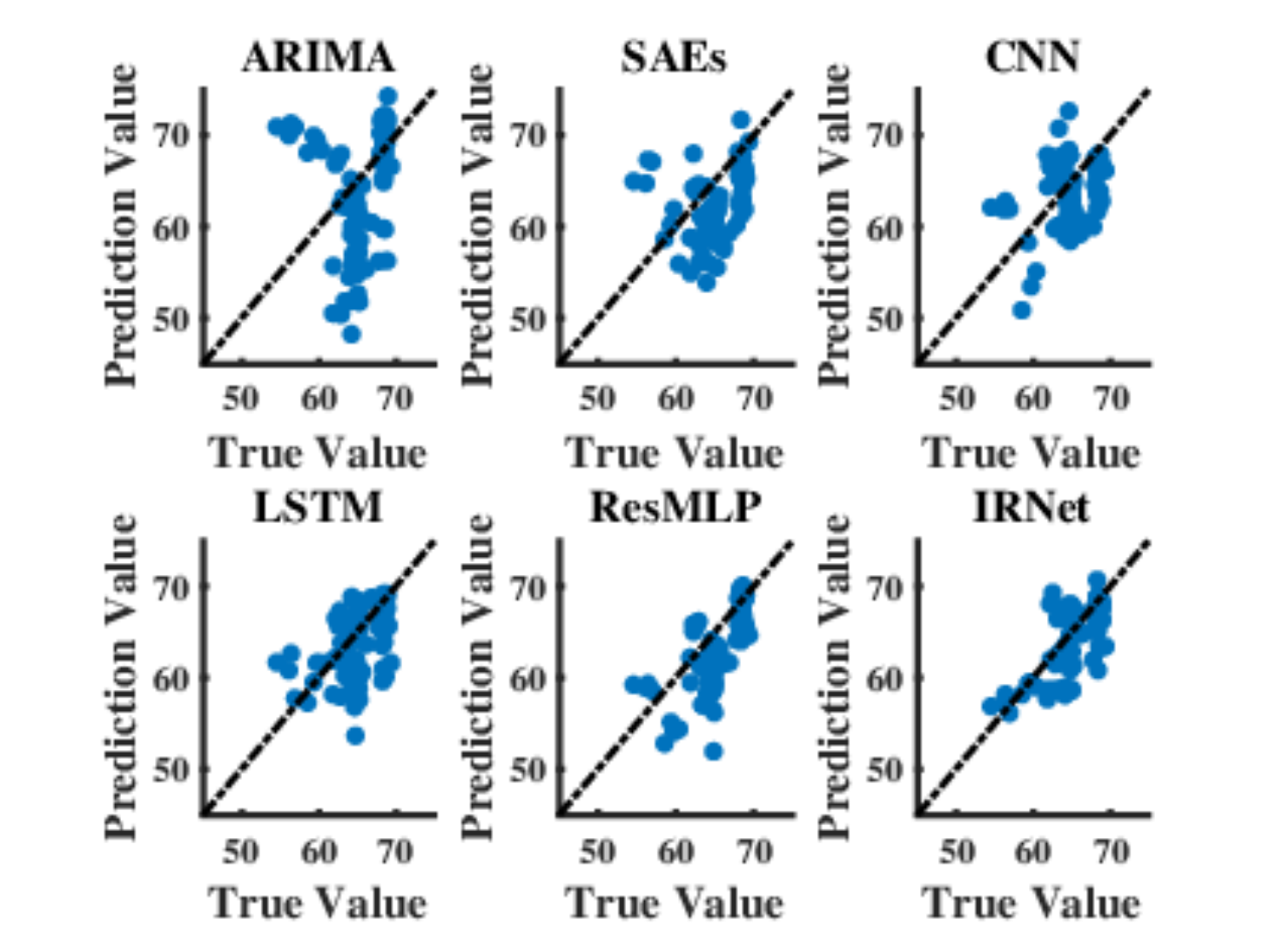}%
\label{fig12a}}
\hfil
\subfloat[]{\includegraphics[width=3.5in]{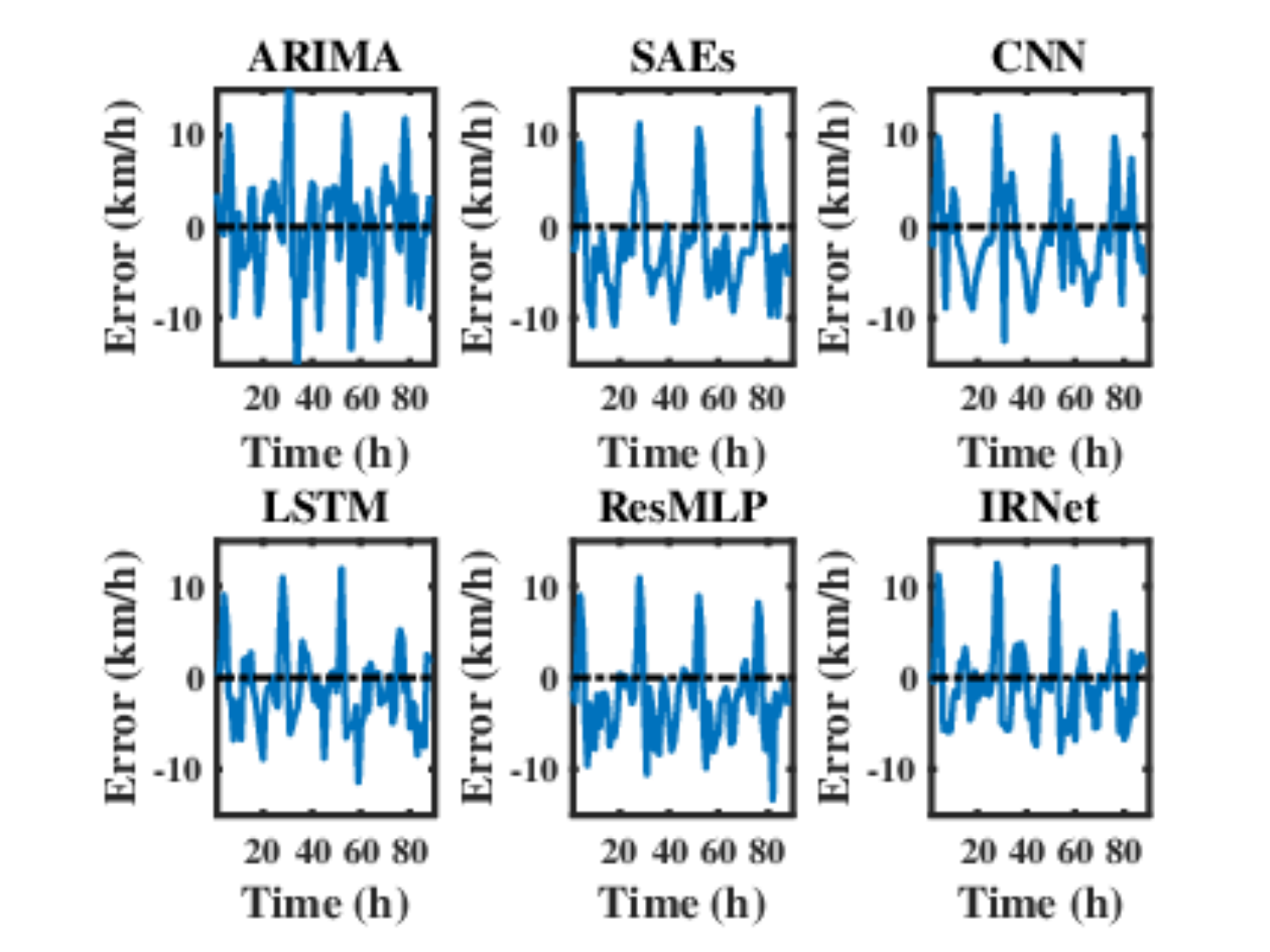}%
\label{fig12b}}
\caption{(a) The scatter plot of true values and predicted values for different methods on the unknown set after fine-tuning. (b) The error plot of true values and predicted values for different methods on the unknown set after fine-tuning.}
\label{fig_12}
\end{figure}

\subsection{Discussion}
The above experiments verify the influence of spatiotemporal information on the prediction ability of the model. The proposed model is compared with the existing road speed prediction models in terms of prediction effect and transfer ability. The conclusion is that the intersection reconstruction model in this paper can reconstruct the intersection to introduce spatial information and strengthen the transfer ability of the model without affecting the prediction effect.
\begin{enumerate}
\item{Larger sequence length and spatial width can provide more temporal and spatial information. It means that more spatiotemporal features can be captured and better prediction results can be obtained. However, in Experiment B, increasing the space width and sequence length led to the reduction of short-term prediction effect and increased computational pressure, so this paper chooses more conservative parameters.}
\item{When the road structure is not processed, the change of the target road results in the structural change of the adjacent roads. Therefore, if the existing road speed prediction model wants to form a pre-trained model suitable for all target roads, it can only use the temporal information of the target road or model the entire road network to introduce spatial information. The intersection reconstruction method proposed in this paper unifies the adjacent road structure of each target road. It solves the problem that the existing road speed prediction models are is difficult to extract spatiotemporal information at the same time. Moreover, in the comparison results of experiment C, it can be seen that the effect of IRNet is slightly ahead of the mainstream time series network LSTM and the more advanced ResMLP. This result proves that the intersection reconstruction method in this paper can improve the prediction effect while simplifying the road structure.}
\item{After fine-tuning the pre-trained model with a small amount of unknown road data, the effect of CNN is greatly deteriorated compared to its own prediction effect. It is because CNN did not extract enough temporal correlations from historical data. During fine-tuning, a small number of unknown roads are used to retrain the regression layer, and the poor effect of the feature extraction layer leads to a serious overfitting phenomenon. The other methods have certain predictive effects after fine-tuning. Among them, IRNet has obtained the best results in multiple sets of experiments, which shows that the introduction of spatial information can greatly improve the transfer effect of the model.}
\end{enumerate}

\section{Conclusions}
This paper propose the IRNet (Transferable Intersection Reconstruction Network) combined with intersection reconstruction method specifically for road speed prediction. After each road is reconstructed into the same structure, the common features of each reconstructed intersection are captured by the convolutional layer, and the temporal correlation and the spatial correlation are obtained through Temporal-LSTM and Spatial-LSTM. Finally, self-attention mechanism is used to combine the spatiotemporal features. Through a series of experiments based on the real dataset PeMS, we prove that IRNet has a better prediction effect than some commonly used baseline models, and has a good level of generalization performance of the model.

In future work, this paper will optimize this work from the following aspects.

First of all, this paper only uses part of the data of one road for training. In theory, after the intersection is reconstructed, the data of multiple roads can be combined into one dataset for training, which will increase the robustness of the model. 

Secondly, the intersection reconstruction method in this paper only compares the correlation between multiple roads at the same intersection and then sorts them. Instead of quantifying the correlation between roads to further process the historical data of adjacent roads, which results in partial distortion of spatial information.

Finally, it is possible to delve deeper into the relationship between the spatiotemporal extent and the prediction span. Using the smallest possible spatiotemporal extent to achieve a better prediction effect can not only save computing resources, but also avoid redundant information interference.
\bibliographystyle{IEEEtran}
\bibliography{ref}

\newpage
\begin{IEEEbiography}[{\includegraphics[width=1in,height=1.25in,clip,keepaspectratio]{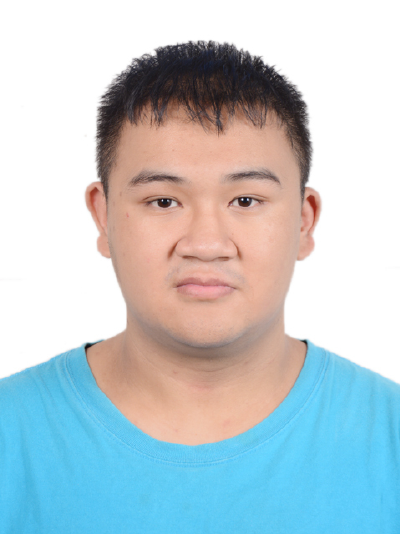}}]{Pengyu Fu} received the B.S. degree in vehicle engineering from Chongqing University, Chongqing, China, in 2019. He is currently working toward the M.S. degree in vehicle engineering from Jilin University, Changchun, China. 

His current research interests include machine learning and neural networks.
\end{IEEEbiography}
\vspace{-10mm}
\begin{IEEEbiography}[{\includegraphics[width=1in,height=1.25in,clip,keepaspectratio]{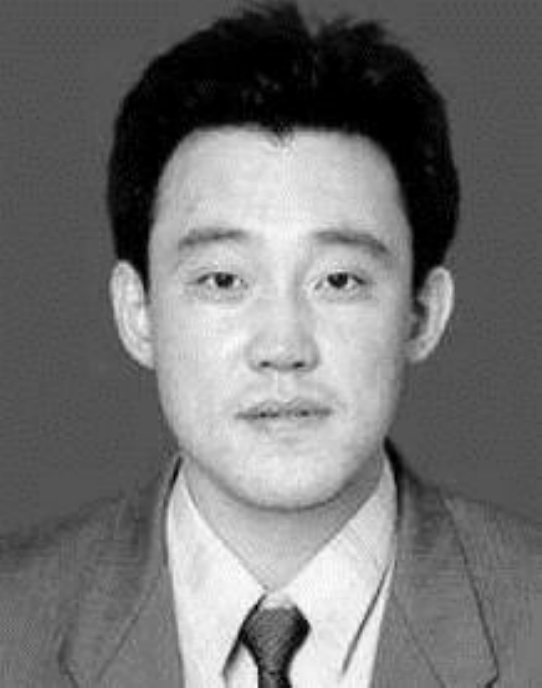}}]{Liang Chu} was born in 1967. He received the B.S., M.S., and Ph.D. degrees in vehicle engineering from Jilin University, Changchun, China. He is currently a Professor and the Doctoral Supervisor with the College of Automotive Engineering, Jilin University. His research interests include the driving and braking theory and control technology for hybrid electric vehicles, which conclude powertrain and brake energy recovery control theory and technology on electric vehicles and hybrid vehicles, theory and technology of hydraulic antilock braking and stability control for passenger cars, and the theory and technology of air brake ABS, and the stability control for commercial vehicle. 

Dr. Chu has been a SAE Member. He was a member at the Teaching Committee of Mechatronics Discipline Committee of China Machinery Industry Education Association in 2006. 
\end{IEEEbiography}
\vspace{-10mm}
\begin{IEEEbiography}[{\includegraphics[width=1in,height=1.25in,clip,keepaspectratio]{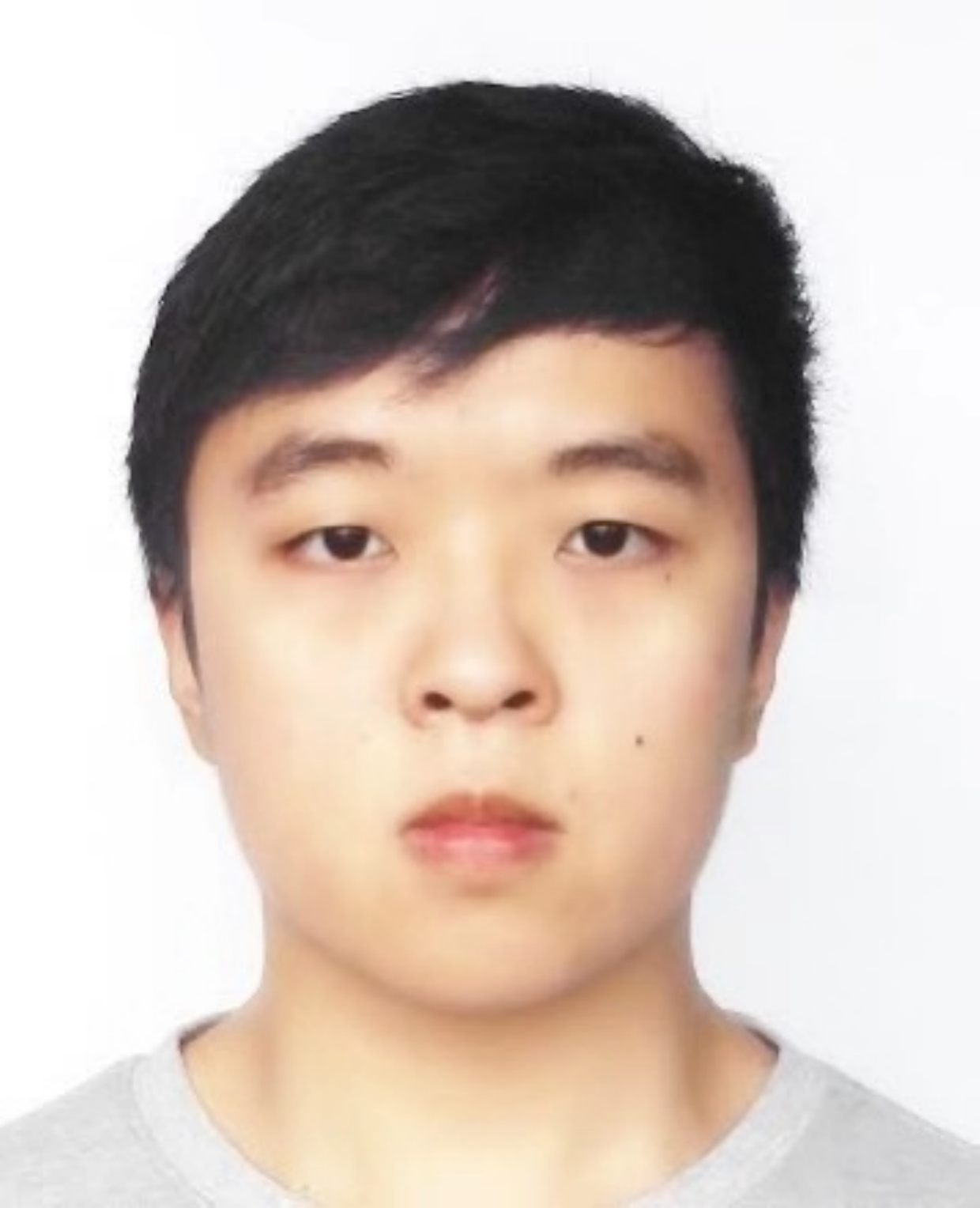}}]{Zhuoran Hou} received the B.S. degree in vehicle engineering from Chongqing University, Chongqing, China, in 2017. and the M.S. in Automotive Engineering from Jilin University, China, in 2020. He is currently pursuing continuous academic program involving doctoral studies in automotive engineering with Jilin University, Changchun, China. 

His research interests include basic machine learning, optimal energy management strategy about plug-in hybrid vehicles. 
\end{IEEEbiography}
\enlargethispage{-9cm}
\newpage
\begin{IEEEbiography}[{\includegraphics[width=1in,height=1.25in,clip,keepaspectratio]{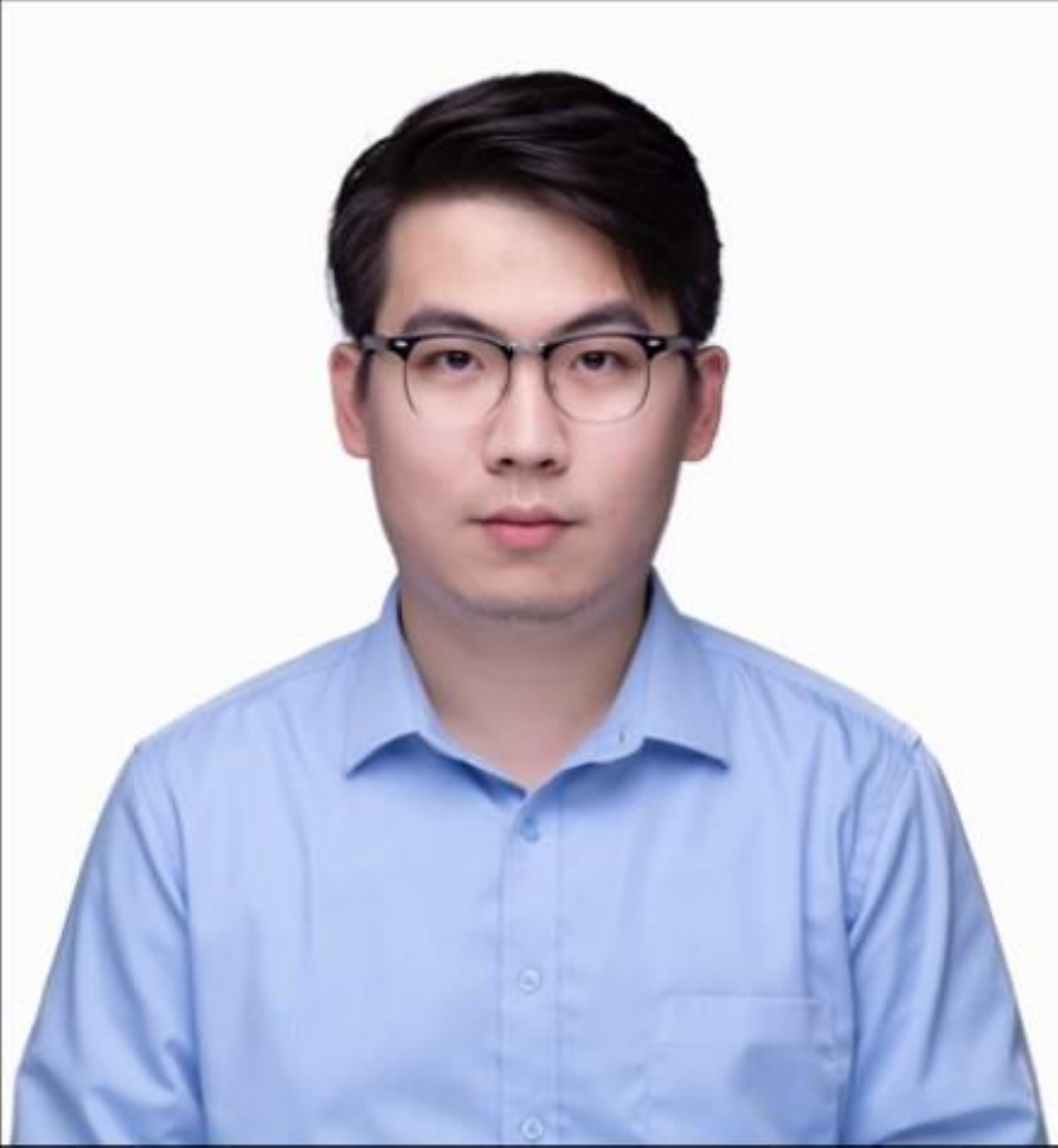}}]{Jincheng Hu} (Student Member, IEEE) received the B.E. degree in information security from the Tianjin University of Technology, Tianjin, China, in 2019 and the M. Sc. degree in information security from the University of Glasgow, Glasgow, UK, in 2022. He is currently working toward the Ph.D. degree in Automotive with the Loughborough university, Loughborough, UK. 

His research interests include reinforcement learning, deep learning, cyber security, and energy management.
\end{IEEEbiography}
\vspace{-10mm}
\begin{IEEEbiography}[{\includegraphics[width=1in,height=1.25in,clip,keepaspectratio]{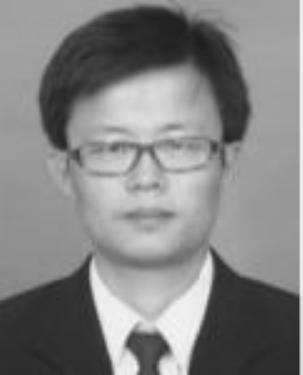}}]{Yanjun Huang} (Member, IEEE) received the Ph.D. degree in mechanical and mechatronics engineering from the University of Waterloo, Waterloo, Canada, in 2016.

He is currently a Professor with the School of Automotive Studies, Tongji University, Shanghai, China. He has authored several books and over 50 papers in journals and conferences. His research interests include the vehicle holistic control in terms of safety, energy saving, and intelligence, including vehicle dynamics and control, hybrid electric vehicle/electric vehicle optimization and control, motion planning and control of connected and autonomous vehicles, and human-machine cooperative driving. Dr. Huang serves as the Associate Editor and Editorial Board Member for the IET Intelligent Transport System, Society of Automotive Engineers (SAE) International Journal of Commercial vehicles, International Journal of Vehicle Information and Communications, Automotive Innovation, etc.
\end{IEEEbiography}
\vspace{-10mm}
\begin{IEEEbiography}[{\includegraphics[width=1in,height=1.25in,clip,keepaspectratio]{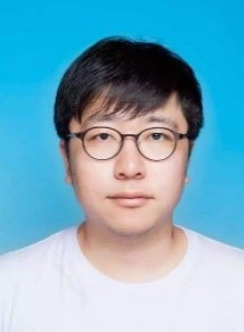}}]{Yuanjian Zhang} (Member, IEEE) received the M.S. in Automotive Engineering from the Coventry University, UK, in 2013, and the Ph.D. in Automotive Engineering from Jilin University, China, in 2018. In 2018, he joined the University of Surrey, Guildford, UK, as a Research Fellow in advanced vehicle control. From 2019 to 2021, he worked in Sir William Wright Technology Centre, Queen’s University Belfast, UK. 

He is currently a Lecturer with the Department of Aeronautical and Automotive Engineering, Loughborough University, Loughborough, U.K. He has authored several books and more than 50 peer-reviewed journal papers and conference proceedings. His current research interests include advanced control on electric vehicle powertrains, vehicle-environment-driver cooperative control, vehicle dynamic control, and intelligent control for driving assist system.
\end{IEEEbiography}
 \enlargethispage{-7.7cm}
\end{document}